\documentclass{article}

\usepackage[utf8]{inputenc}
\usepackage[colorlinks = true,
            linkcolor = blue,
            urlcolor  = blue,
            citecolor = blue,
            anchorcolor = blue]{hyperref}
\usepackage{amsmath,amssymb,amsfonts,amsthm,bm}
\usepackage{color,graphicx}
\usepackage[a4paper, total={6.5in,8.5in}]{geometry}
\usepackage{adjustbox}

\usepackage{amssymb}
\usepackage{empheq}
\usepackage{mathtools}
\usepackage{graphicx}
\usepackage{bbm}
\usepackage{hyperref}
\usepackage{dirtytalk}
\usepackage{stmaryrd}
\usepackage{fancyhdr}
\usepackage{cleveref}
\usepackage{microtype}
\usepackage{graphicx}
\usepackage{subcaption}
\usepackage{booktabs}
\usepackage{makecell}
\usepackage{caption}
\usepackage{amsmath}
\usepackage{booktabs} 
\usepackage{tablefootnote}
\usepackage{footnote}
\usepackage{hyperref}
\usepackage{tabularx}
\usepackage[ruled,vlined]{algorithm2e}
\usepackage{epigraph}

\usepackage{verbatim}
\usepackage{listings}
\usepackage{algpseudocode}
\usepackage[affil-it]{authblk} 
\usepackage{textcomp}
\usepackage{multirow}
\usepackage[toc,page]{appendix}
\usepackage{xspace}
\usepackage{wrapfig}
\usepackage[numbers, sort, comma, square]{natbib}
\usepackage{float}
\usepackage{todonotes}

\usepackage{macros}

\newif\ifshownotes
\shownotestrue

\newif\ifarxiv
\arxivfalse

\usepackage{longtable}
\date{}
\author[]{Piersilvio De Bartolomeis$^1$}
\author[]{Jacob Clarysse$^1$}
\author[]{Amartya Sanyal$^2$}
\author[]{Fanny Yang$^1$}
\affil[]{$^1$Department of Computer Science, ETH Zürich\\%
    $^2$Max Planck Institute for Intelligent Systems, T\"ubingen
}

\title{How robust accuracy suffers from certified training \\with convex relaxations}

\begin{document}

\maketitle
\setcounter{page}{1}

\begin{abstract}
Adversarial attacks pose significant threats to deploying state-of-the-art classifiers in safety-critical applications. Two classes of methods have emerged to address this issue: empirical defences and certified defences. Although certified defences come with robustness guarantees, empirical defences such as adversarial training enjoy much higher popularity among practitioners. In this paper, we systematically compare the standard and robust  error of these two robust training paradigms across multiple computer vision tasks. We show that in most tasks and for both \(\ell_\infty\)-ball and \(\ell_2\)-ball threat models, certified training with convex relaxations suffers from worse standard and robust error than adversarial training. We further explore how the error gap between certified and adversarial training depends on the threat model and the data distribution. In particular, besides the perturbation budget, we identify as important factors the shape of the perturbation set and the implicit margin of the data distribution. We support our arguments with extensive ablations on both synthetic and image datasets.
\end{abstract}

\section{Introduction}
State-of-the-art classifiers are known to be vulnerable to adversarial perturbations of the input. These perturbations, whether perceptible or imperceptible, can manipulate the input in such a way as to drastically alter the classifier's predictions~\citep{biggio_evasion_2013,szegedy_intriguing_2014}. Hence, robustness to such \emph{adversarial attacks} has become a crucial design goal when deploying machine learning models in safety-critical applications. 

 More precisely, for any given distribution \(\mathcal{D}\) and loss function $L$,
we aim to find a classifier $\nn:\RR^\di \to \RR^k$
parameterised by $\theta \in \RR^p$
that minimises the \emph{robust loss}
\begin{align}
\label{eq:robust_opt}
\min_\theta \rrisk(\theta)   \; \;  \text{where}  \;\; \rrisk(\theta)  \defn \EE_{(x,y) \sim \mathcal{D}} \left[\max_{\delta \in \BB_{\epsilon}} L(\nn(x + \delta),y)\right],
\end{align}
the threat model \(\BB_{\epsilon} \defn \{ \delta : \norm{\delta}_p \leq \epsilon
\}\) is a  bounded $\ell_p$-ball centred at the origin, and \(\rrisk(\theta)\) denotes the robust error when \(L\) is the 0-1 loss. 
The main challenge in solving the optimisation problem presented in~\Cref{eq:robust_opt} is that, when $\theta$ parameterises a neural network, the inner maximisation becomes a non-convex optimisation problem and computationally intractable~\citep{katz_reluplex_2017,weng_towards_2018}. In order to overcome the computational barrier, two methods of approximation have been widely discussed in the literature so far:
\emph{empirical} defences such as adversarial training 
and \emph{certified} defences, including randomised smoothing and convex relaxations. 


Adversarial training
(\madry)~\citep{goodfellow_explaining_2015, madry_towards_2018} is one
of the most popular empirical defences to date: it minimises the
empirical robust loss in~\Cref{eq:robust_opt}  by approximately
solving the inner maximisation with iterative first-order 
optimisation methods. Although adversarial training is favoured for its simplicity and computational efficiency, it lacks the robustness guarantees that are essential in 
safety-critical applications. In particular, 
 even though empirical evaluation tools like AutoAttack~\citep{croce_reliable_2020} can provide non-trivial lower bounds of the robust error, it is worth noting that these lower bounds might substantially underestimate the true robust error.

To address this limitation, certified defences train 
neural networks for which it is possible to  
obtain non-trivial upper bounds of the robust error. In this paper, we focus our attention on \emph{certified training}, i.e. the widely studied class of certified defences based on convex relaxations.  The key idea
underlying certified training methods is to solve a convex relaxation of the
optimisation problem in~\Cref{eq:robust_opt}, by replacing
the non-convex ReLU constraint sets with larger convex sets~\citep{wong_provable_2018, raghunathan_certified_2018,dvijotham_dual_2018}.

\begin{figure*}[!t]
 \begin{subfigure}[b]{0.48\textwidth}
      \centering
\includegraphics[scale=0.33]{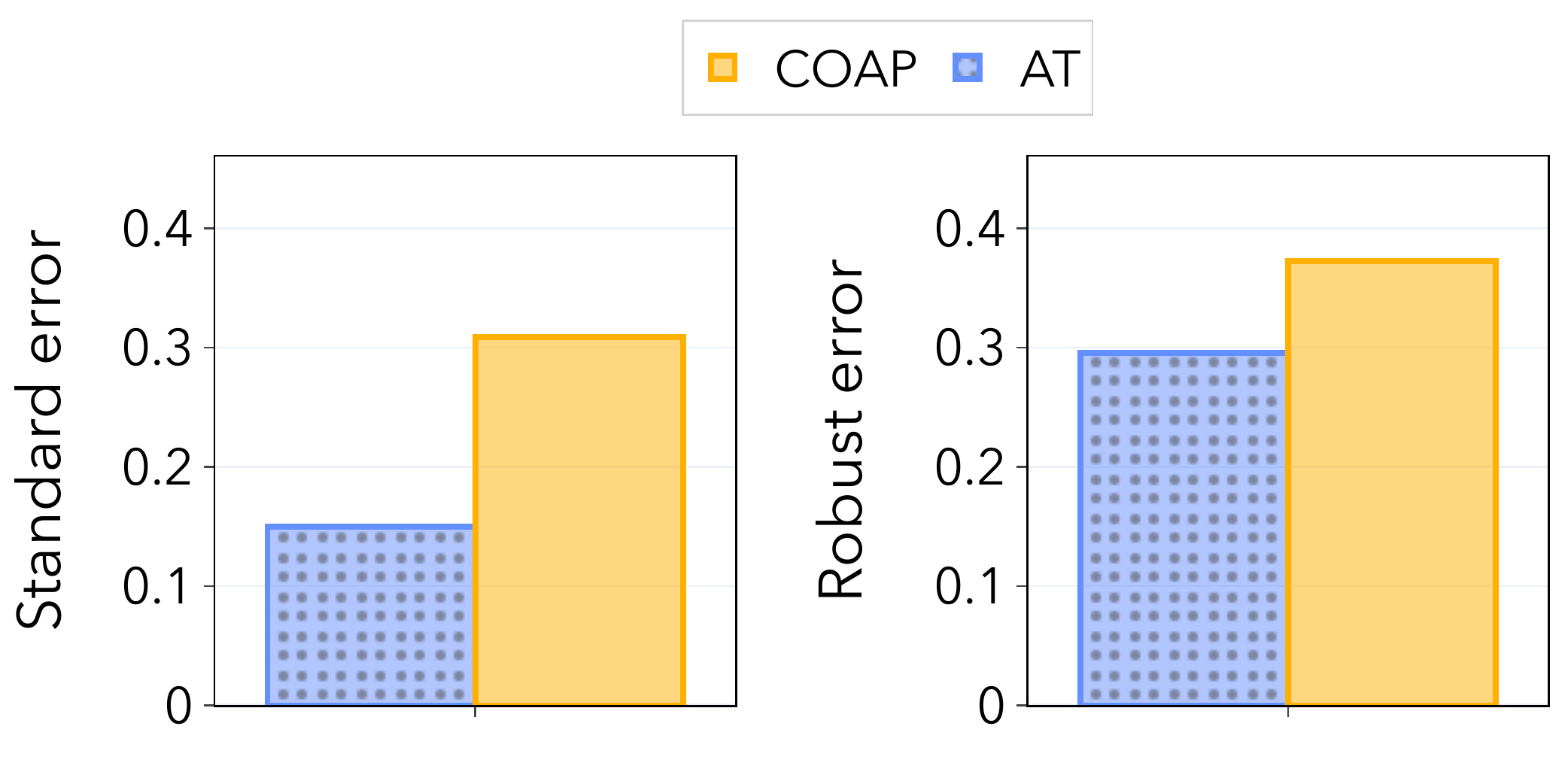}
        \caption{ $\ell_\infty$-ball }\label{fig:cifar_linf}
     \end{subfigure}
     \centering
      \hfill
     \begin{subfigure}[b]{0.48\textwidth}
         \centering
\includegraphics[scale=0.33]{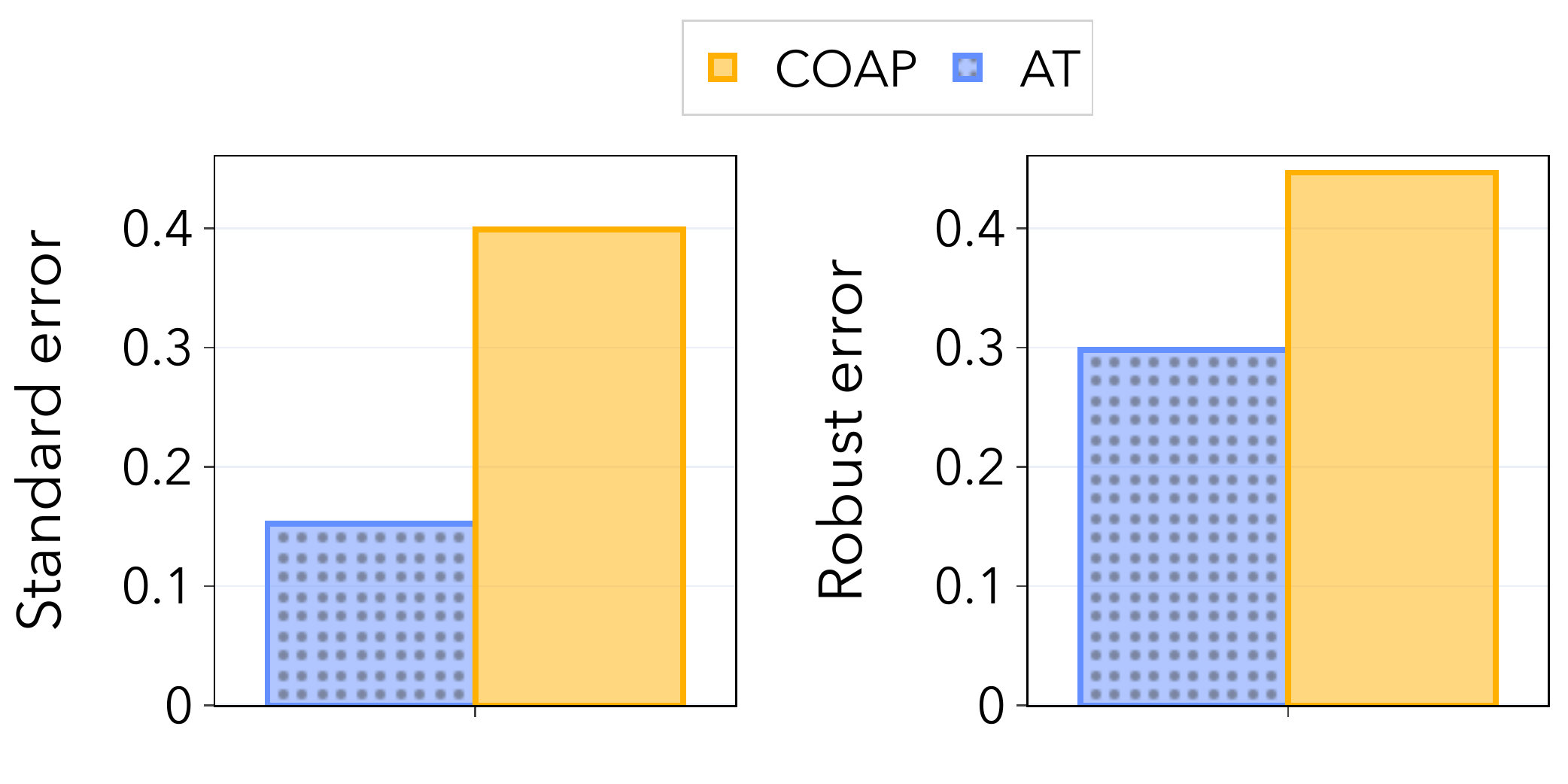}
         \caption{$\ell_2$-ball  }
 \label{fig:cifar_l2}
    \end{subfigure}
                \caption{Standard and robust error  of adversarial (dotted bars) and certified training (solid bars) on the CIFAR-10 
            test set. Models were trained for robustness against: (a) \(\ell_\infty\)-ball perturbations with radius \(\epsilon_\infty=1/255\), and (b) \(\ell_2\)-ball perturbations with radius \(\epsilon_2=36/255\). We report the best performing certified training method among many convex relaxations~(FAST-IBP~\citep{shi2021fast}, IBP~\citep{gowal_scalable_2019}, \zhang~\citep{zhang_towards_2020,xu_automatic_2020} and \wong~\citep{wong_scaling_2018, wong_provable_2018}). 
            We refer the reader to~\Cref{sec:images_evaluation}
            for further details on the models and robust evaluation. }
           \label{fig:cifar_experiments}
\end{figure*}

While certified training offers robustness guarantees, adversarial training is frequently the preferred choice among practitioners. 
Beyond computational efficiency issues, this preference stems from the suboptimal standard and robust error achieved with this class of certified defences~\citep{mao2023taps}. As we discuss in Section~\ref{sec:related}, so far, the drawbacks of certified training have received considerably less attention in the literature than other types of defences, such as randomised smoothing~\citep{mohapatra_hidden_2021,avidan_spectral_2022, blum_random_2020, kumar_curse_2020}. Further, the
most prominent works on certified training ~\citep{wong_provable_2018, zhang_towards_2020, xu_automatic_2020, gowal_scalable_2019, shi2021fast, muller2023certified}
do not include a direct comparison of their methods with adversarial training, nor are there any existing explanations for the error gap between these two paradigms of robust training.

In this paper, we fill this void in the literature and systematically compare the robust and standard error of certified and adversarial training across three widely adopted computer vision datasets.
More specifically, our contributions can be summarised  as follows:

\begin{itemize}
\item 	In~\Cref{sec:images_evaluation}, we show that, in most tasks and for both \(\ell_\infty\)-ball and \(\ell_2\)-ball threat models, certified training suffers from worse standard and robust error than adversarial training (e.g. see~\Cref{fig:cifar_linf,fig:cifar_l2}). 
 	\item 
In~\Cref{sec:factors}, we explore how the error gap between certified and adversarial training depends on the \renewcommand{\labelenumi}{(\roman{enumi})}
 \begin{enumerate} 	
 \item \textbf{Threat model} --  
 via the perturbation budget and the shape of the perturbation set.  
 \item \textbf{Data distribution} --  via the implicit margin of the data, which denotes the minimum distance in feature space between any two data points from different classes.
 \end{enumerate}
 
  
  \item In~\Cref{sec:neurons}, we propose a possible explanation for the error gap between certified and adversarial training. Specifically, through a series of ablation studies and illustrations, we show that the above factors influence the number of \emph{unstable neurons}, which in turn affects the error gap. 

  
\end{itemize} 

\section{Systematic comparison between adversarial and certified training}
\label{sec:images_evaluation}

In this section, we systematically compare the standard and robust error of adversarial  and certified training, for both \(\ell_2\)-ball and \(\ell_\infty\)-ball threat models. We consider three widely adopted computer vision datasets: Tiny
 ImageNet~\citep{le_tiny_2015},
CIFAR-10~\citep{krizhevsky_learningmultiple_2009} and
 MNIST~\citep{lecun_gradient-based_1998}. We note that certified training with convex relaxations does not scale to larger datasets such as ImageNet, hence we omit them from our comparison.

 \subsection{Experimental setup}
 \paragraph{Adversarial defences} 
 Among certified training methods, we focus on the most popular ones and cover a wide range of convex relaxations techniques. In particular, we consider:
 \begin{itemize}
 	\item Convex outer adversarial polytope
 (\wong)~\citep{wong_scaling_2018, wong_provable_2018}, which uses the DeepZ relaxation~\citep{singh_fast_2018} and achieves state-of-the-art certified robustness under \(\ell_2\)-ball
 perturbations.
 \item Interval bound propagation (IBP)~\citep{gowal_scalable_2019}, which uses the Box relaxation~\citep{mirman_differentiable_2018}.
 \item Fast IBP~\citep{shi2021fast}, which is a computationally
 more efficient version of IBP and achieves state-of-the-art certified robustness under $\ell_\infty$-ball perturbations. 
 \item
\zhang~\citep{xu_automatic_2020}, which combines
 the tight convex relaxation CROWN~\citep{zhang_efficient_2018} with
 IBP.
 \end{itemize}
  We compare the certified training methods against the
 most popular empirical defence to date, adversarial training
 (\madry)~\citep{goodfellow_explaining_2015, madry_towards_2018},
 which is the go-to approach for adversarial robustness among
 practitioners.
  \paragraph{Models and evaluation} For CIFAR-10, we  train a
 residual network~(ResNet) and for MNIST we train a vanilla convolutional
 neural network~(CNN). Both architectures were introduced in
 \citet{wong_scaling_2018} as standard benchmarks for certified training. For Tiny ImageNet, we train the WideResNet introduced in~\citet{xu_automatic_2020}. We refer the reader to \Cref{apx:image_experiments} for complete experimental
 details.

We evaluate the performance of these models according to standard and robust error. More precisely, standard error is defined as in~\Cref{eq:robust_opt} by setting $L$ to the 0-1 loss function and \(\epsilon=0\).
 As an approximation for the population quantity, we compute the empirical standard and robust error on the test set. Further, since exact evaluation of the robust error is computationally infeasible, we evaluate the models with AutoAttack
 (AA+)~\citep{croce_reliable_2020}, which is widely considered to be one of the most reliable empirical evaluation tools to date. 

 
Our threat models in this section correspond to $\ell_2$-ball and $\ell_\infty$-ball perturbations. For the sake of clarity, we present our experimental results with a single perturbation budget for each dataset and threat model, as reported in~\Cref{table:image_experiments}~(caption). Nevertheless, we provide additional experimental results with a wide range of perturbation budgets in~\Cref{apx:complete_comparison}. Note that we intentionally choose small perturbation budgets for \(\ell_2\)-ball perturbations, as larger budgets result in trivial standard and robust error for certified training.


\subsection{Standard and robust error gap} 
We present the results of our comparison in~\Cref{table:image_experiments}. Generally,  for both \(\ell_2\)-ball and \(\ell_\infty\)-ball threat models, we observe that certified training suffers from worse standard and robust error than adversarial training. An interesting exception to this pattern is found with the MNIST dataset and the \(\ell_\infty\)-ball threat model, where the best convex relaxation achieves smaller robust error than adversarial training.

\paragraph{Discussion} For MNIST, in~\Cref{table:image_experiments} (first row) we observe a gap in standard error of up to $2\%$ between adversarial and certified training, under both \(\ell_2\)-ball and \(\ell_\infty\)-ball perturbations. Similarly, for robust error under \(\ell_2\)-ball perturbations, we observe a $6\%$ gap between the best convex relaxation and \madry. However, under \(\ell_\infty\)-ball perturbations the best convex relaxation achieves lower robust error than \madry. 

For CIFAR-10, in~\Cref{table:image_experiments} (second row) we observe a standard error gap of up to \(25\%\) between the best convex relaxation and \madry, under both \(\ell_2\)-ball and \(\ell_\infty\)-ball perturbations. Unlike MNIST, for robust error, we observe a gap of up to \(15\%\), under both threat models. 

For Tiny ImageNet, in~\Cref{table:image_experiments} (third row) we observe a significant standard error gap of up to $35\%$ between the best convex relaxation and \madry, under both \(\ell_2\)-ball and \(\ell_\infty\)-ball perturbations. Similarly, we observe a significant robust error gap for both threat models, of up to $20\%$. Note that \wong does not scale to Tiny ImageNet, and hence we omit it from our comparison.

We observe that even though \wong is not traditionally recognised as a state-of-the-art certified training method under \(\ell_\infty\)-ball perturbations, given its subpar performance for larger perturbation budgets, it surprisingly surpasses all other convex relaxations for smaller perturbation budgets on the CIFAR-10 test set. Additionally, it generally performs better under \(\ell_2\)-ball perturbations for both MNIST and CIFAR, marking it as an effective convex relaxation in these settings.

\begin{table}[t!]
        \centering
        \begin{tabular}{|c|c|c|}
            \hline
            & $\ell_{\infty}$-ball & $\ell_{2}$-ball \\ 
            \hline
            \multirow{2}{*}[10ex]{\rotatebox[origin=c]{90}{MNIST}} & \includegraphics[scale=0.3]{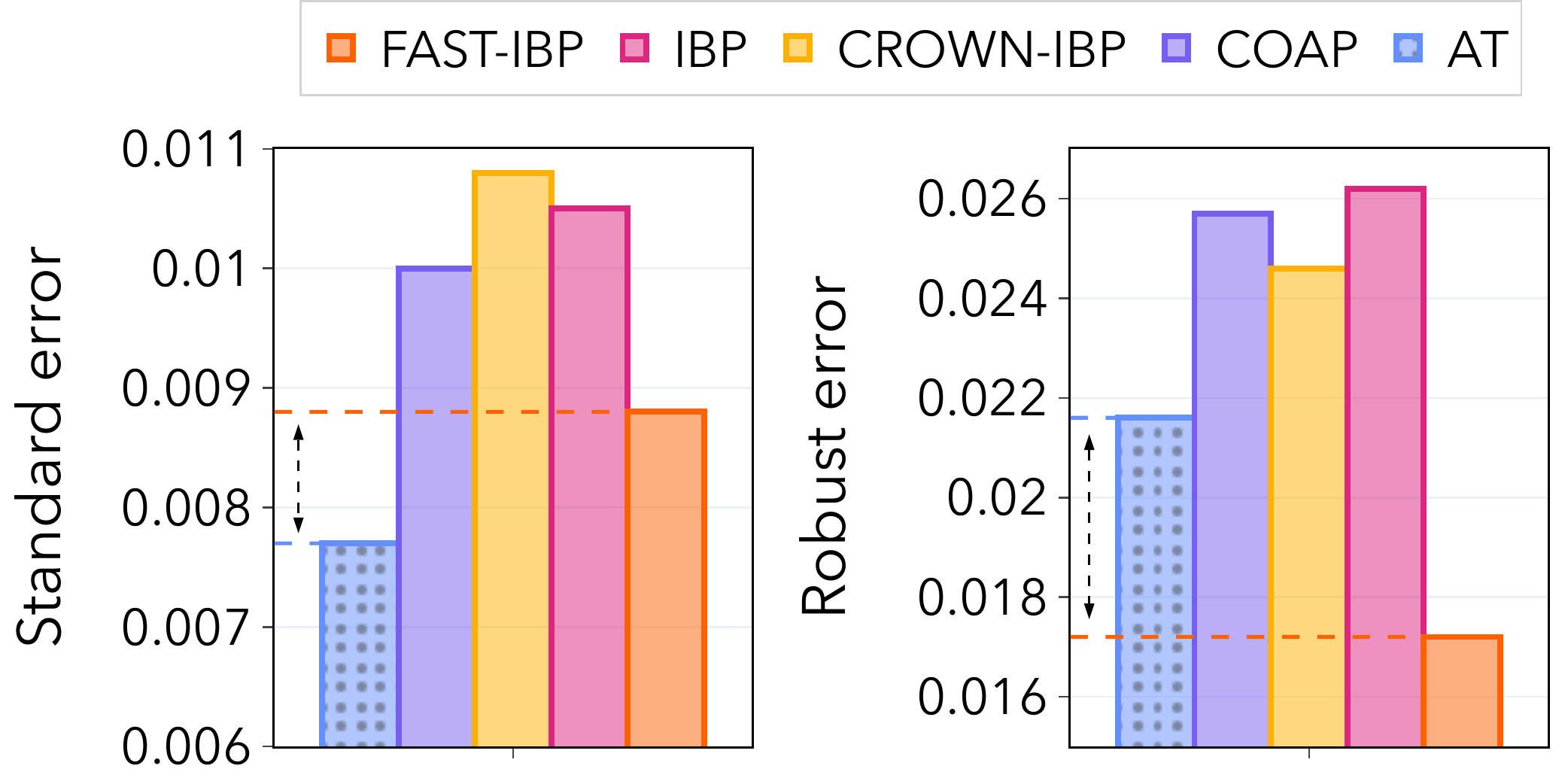} & \includegraphics[scale=0.3]{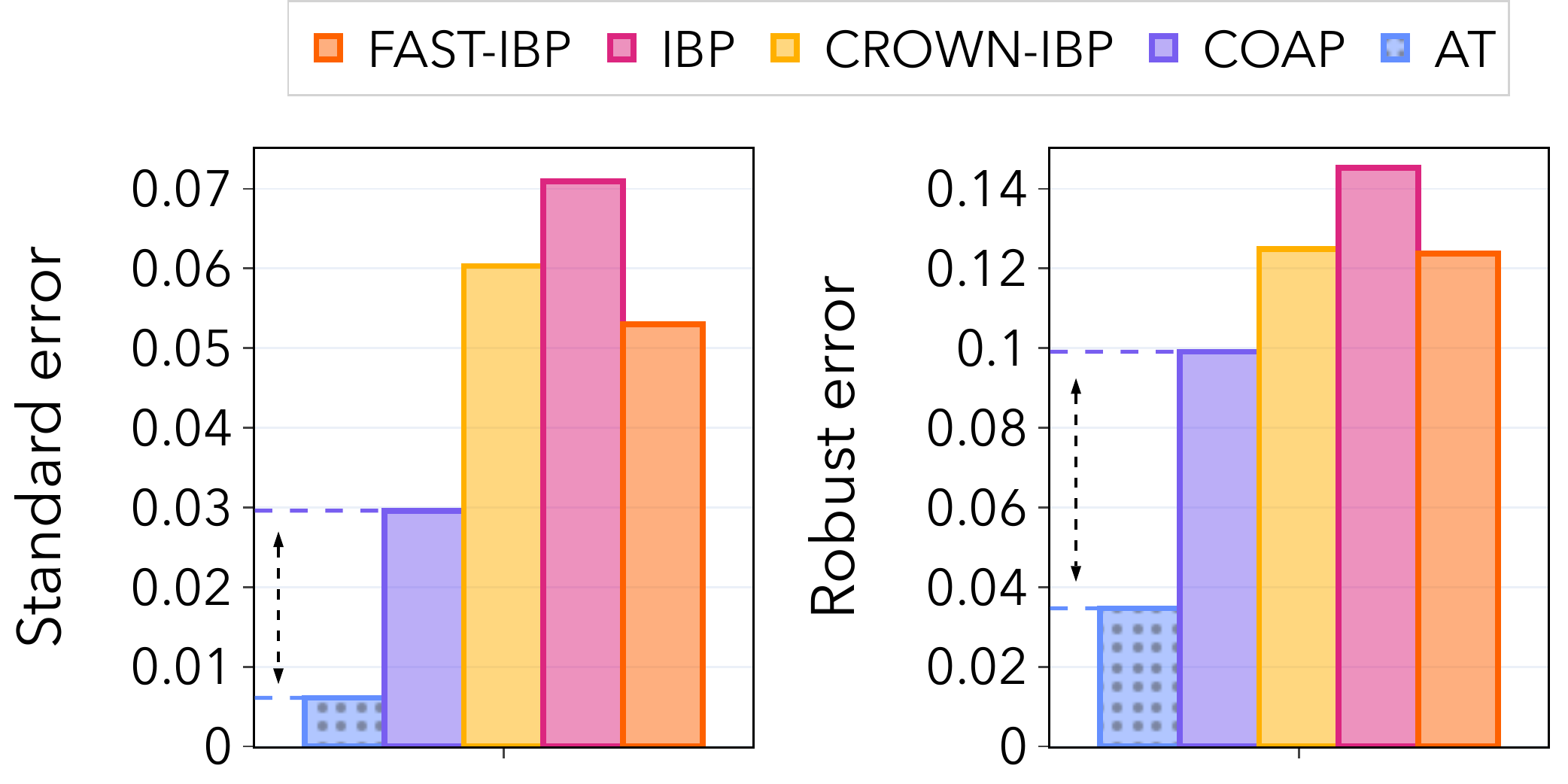} \\
            & & \\
            \hline
            \multirow{2}{*}[11ex]{\rotatebox[origin=c]{90}{CIFAR-10}} & \includegraphics[scale=0.3]{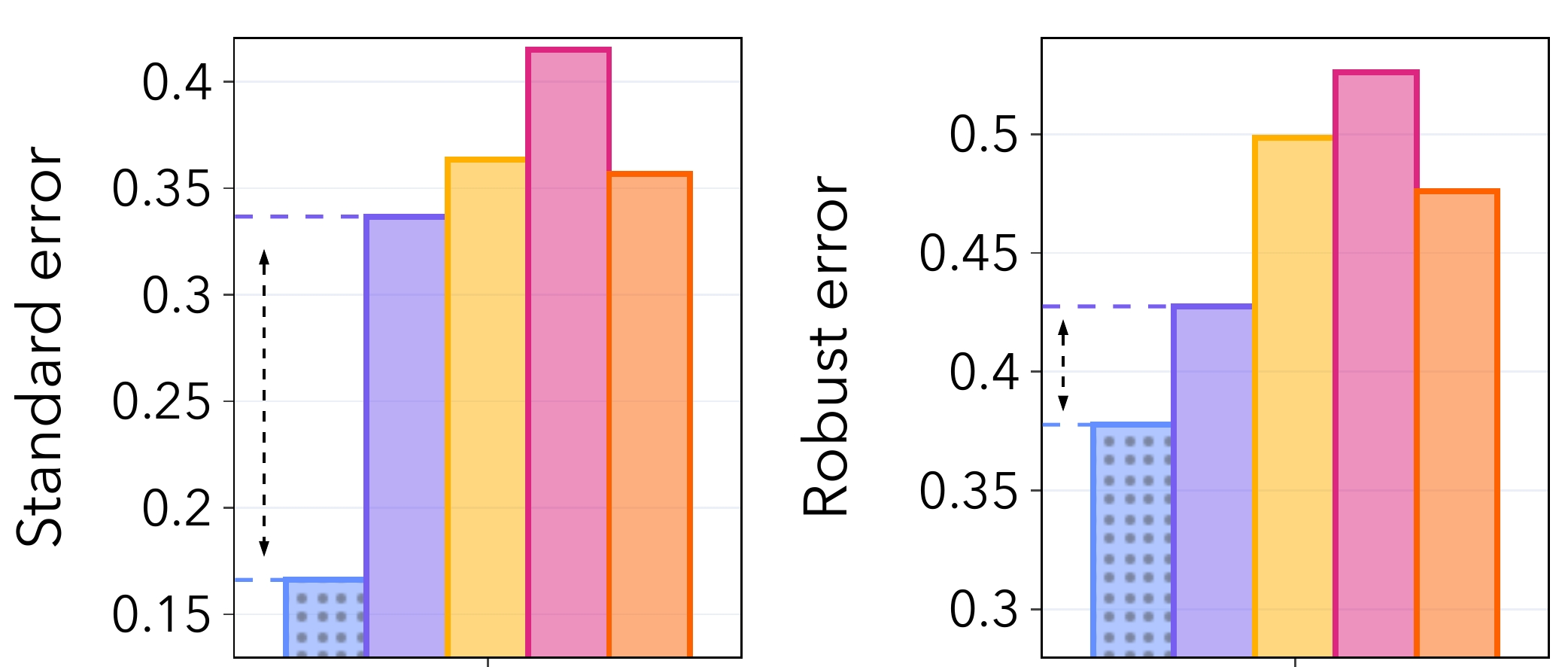} & \includegraphics[scale=0.3]{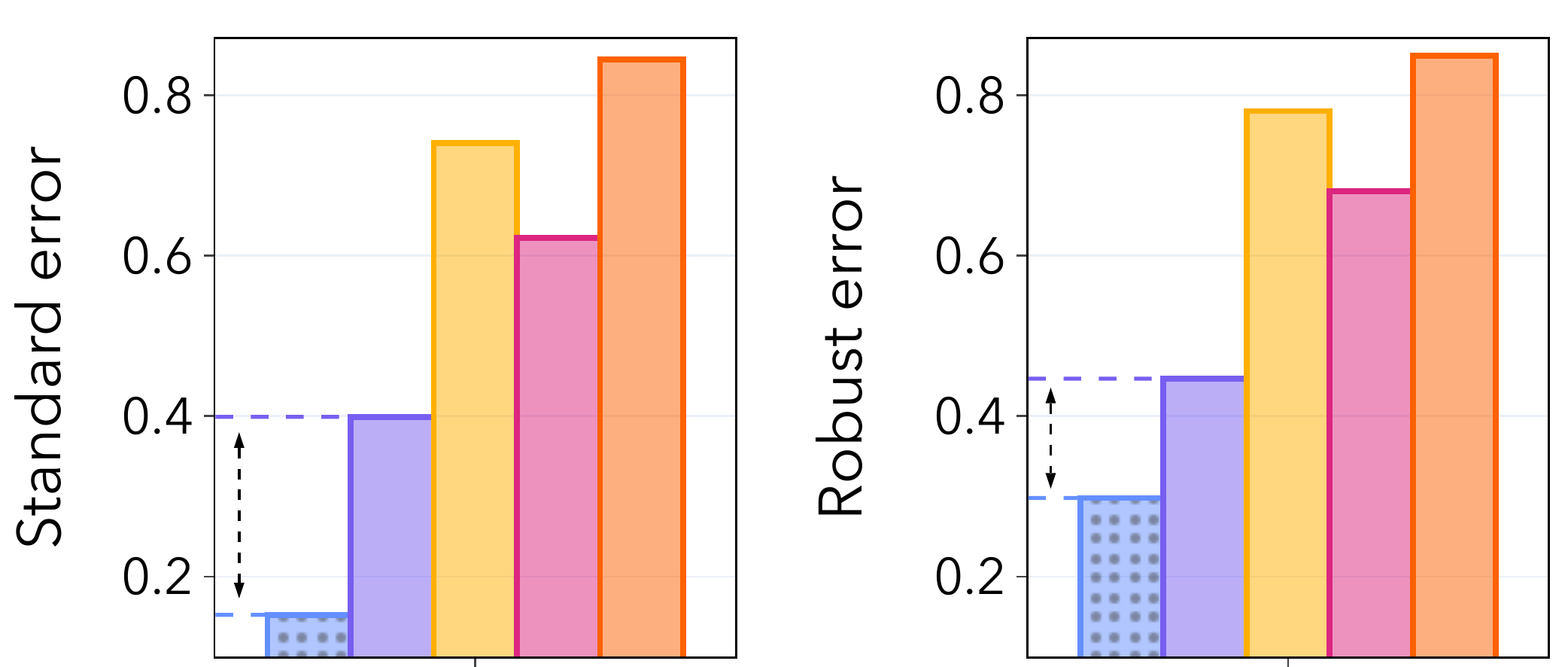} \\
            & & \\
            \hline
            \multirow{2}{*}[12ex]{\rotatebox[origin=c]{90}{Tiny ImageNet}} & \includegraphics[scale=0.3]{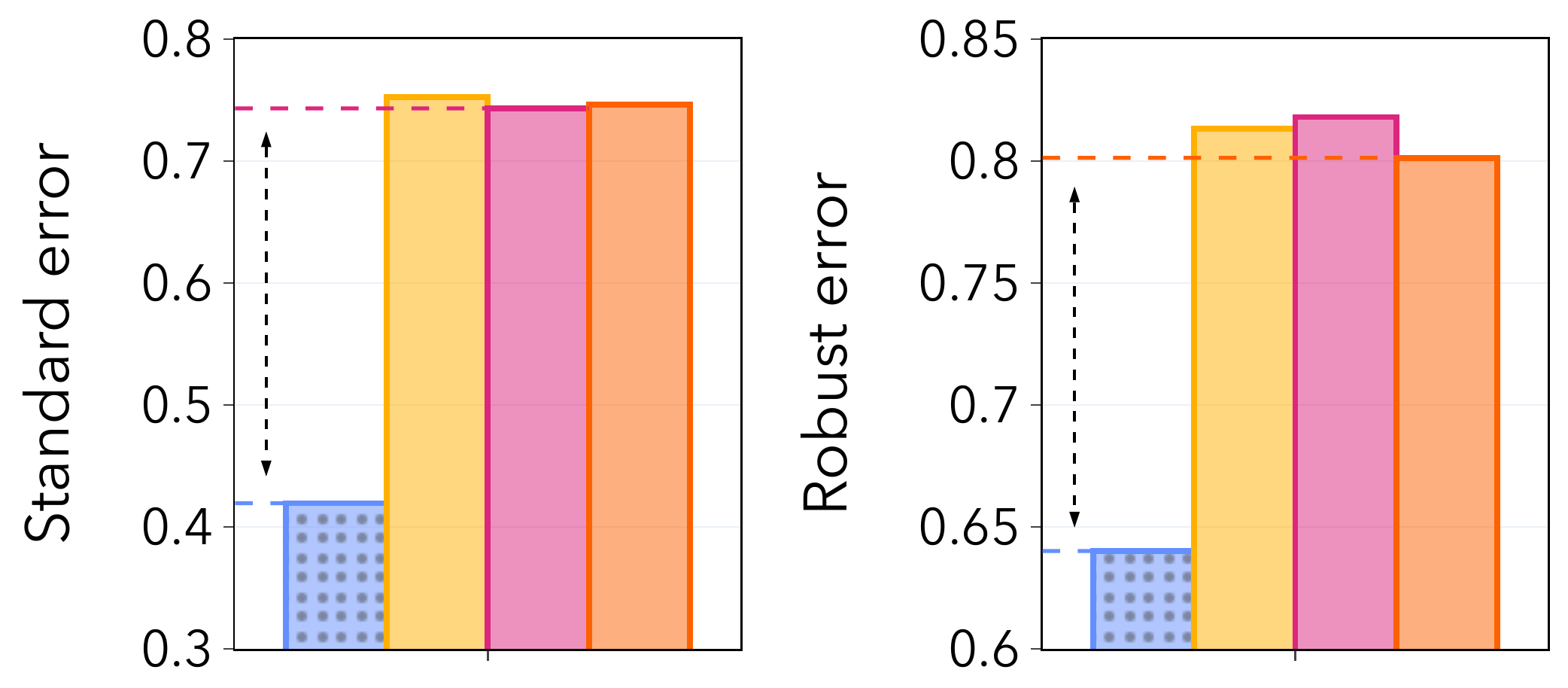} & \includegraphics[scale=0.3]{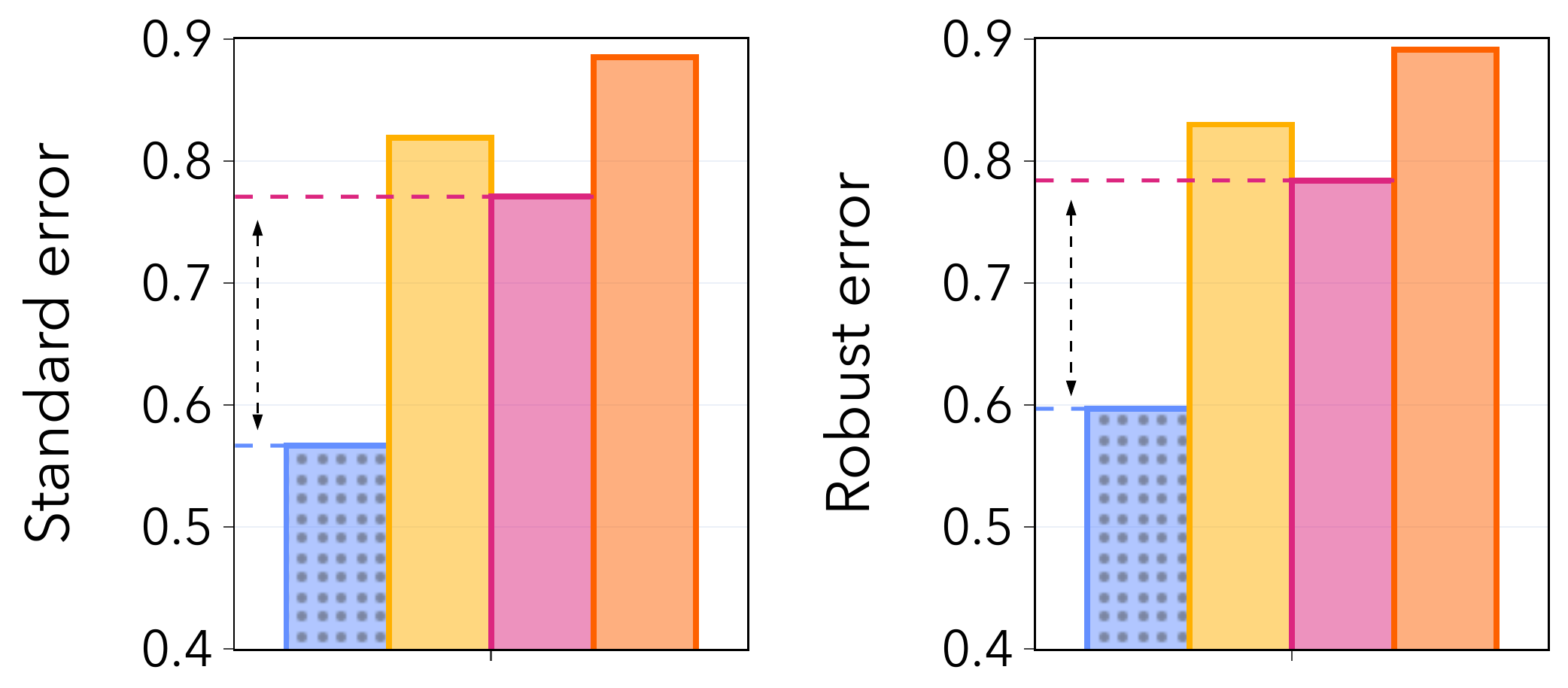}\\
            & & \\
            \hline
        \end{tabular}
        \vspace{2mm}
                \caption{Standard and robust error of adversarial  (dotted bar) and certified training (solid bar, tightest to loosest convex relaxation from left to right) on the MNIST ($\epsilon_\infty=0.1, \epsilon_2=0.75)$, CIFAR-10 ($\epsilon_\infty=2/255, \epsilon_2=36/255$) and Tiny ImageNet ($\epsilon_\infty=1/255, \epsilon_2=5/255$) test sets. 
 }
      \label{table:image_experiments} 

    \end{table}

\paragraph{Tight convex relaxations are better for \(\ell_2\)-ball perturbations} We observe a significantly more pronounced standard and robust error gap
for \(\ell_2\)-ball perturbations compared to \(\ell_\infty\)-ball perturbations.  Surprisingly, we
find that the paradox of certified
training~\citep{jovanovic_paradox_2021} — the notion that loose
interval-based training often yields better performance than tighter
relaxations — does not hold for \(\ell_2\)-ball perturbations and for \(\ell_\infty\)-ball perturbations when the perturbation budget is small (e.g. CIFAR-10 with \(\epsilon_\infty=\frac{2}{255}\)). Instead, we notice that tighter convex relaxations actually enhance
performance in this setting. Notably, COAP~\citep{wong_provable_2018,
wong_scaling_2018}, which is the tightest among the convex relaxations
considered, stands out as the best-performing method against
\(\ell_2\)-ball perturbations on both MNIST and CIFAR-10 test sets. Conversely, IBP and FAST-IBP, which are the loosest
convex relaxation considered, emerge as the least effective.


\section{Which factors influence the error gap?}
\label{sec:factors}

In this section, we explore how the standard and robust error gap between certified and adversarial training depends on the threat model and  the data distribution. 
Besides the perturbation budget $\epsilon$, we find that the shape of the perturbation set plays a crucial role: the more \emph{aligned} it is with the shortest path towards the (robust Bayes optimal) decision boundary, the larger the error gap. Furthermore, the natural distribution of the data also affects the phenomenon via the \emph{implicit margin} $\gamma$: a small minimum distance between two classes in feature space leads to a large  error gap. We discuss the intuition underlying these factors in~\Cref{sec:neurons}.


 
 \subsection{Experimental setup}

  

\paragraph{Datasets} We use one synthetic dataset and one image dataset to study the effect of the aforementioned factors on the error gap.
For our controlled synthetic setting, we consider the concentric spheres distribution studied in~\citet{gilmer_adversarial_2018, nagarajan_uniform_2019}. 
To sample from the concentric spheres distribution with radii $0<R_0<R_1$, we first draw a binary label $y\in \{0, 1\}$ with equal probability, and then a covariate vector $x \in R^d$ uniformly from the sphere of radius $R_{y}$. Note that the implicit margin of the concentric spheres distribution is given by $\gamma \vcentcolon=R_1-R_0$. For image datasets, we present our ablations on CIFAR-10 but observe similar trends on MNIST and provide additional experiments in~\Cref{apx:mnist_ablations}.

\paragraph{Models and robust evaluation}
For all CIFAR-10 experiments, we  train a
vanilla convolutional neural network (CNN) as specified in
\citet{wong_scaling_2018}. For all  concentric spheres experiments, we train a multilayer perceptron with $W=100$ neurons and one hidden layer. For simplicity of exposition, we compare adversarial training with \wong throughout this section. We choose \wong as a representative for certified training since it outperforms the other convex relaxations under $\ell_2$-ball perturbations. We refer the reader to~\Cref{apx:exp_details} for complete experimental details.
 
For the ablations on the perturbation budget and the margin, we specifically concentrate on \(\ell_2\)-ball perturbations, as the phenomena we aim to investigate are more prominent in this context. Nevertheless, we observe similar trends for \(\ell_\infty\)-ball perturbations, and we provide additional experiments in~\Cref{apx:linf_ablations}. As for the previous section, we evaluate the empirical robust error  for \(\ell_2\)-ball perturbations using AutoAttack
 (AA+)~\citep{croce_reliable_2020}. 

\subsection{Factor (i): Shape of the perturbation set}
\label{sec:fac1}

The first factor we investigate is the shape of the perturbation set. 
In particular, we study the alignment of the perturbation set with the shortest path to the (robust Bayes optimal) decision boundary, which we call the \emph{signal direction}.

More formally, we define the signal direction \(s(x,y)\) for a data point \((x,y) \in \mathcal{X} \times \mathcal{Y}\) and the robust Bayes optimal classifier \(f^*: \mathcal{X} \to \mathcal{Y}\), as the direction along the shortest path to the decision boundary

\begin{align*}
    s(x, y) \defn \underset{\delta: \|\delta\|_2=1}{\mathrm{argmin}} \left\{ \underset{\epsilon \geq 0}{\mathrm{min}} \; \epsilon \; : \; f^{*}(x+\epsilon \cdot \delta) \neq y \right\}.
\end{align*}


Then, given a data distribution $\mathcal{D}$ 
and a threat model \(\BB_\epsilon\), we define the alignment of the perturbation set as 
\begin{align*}
    \alignment(\BB_\epsilon)  &\defn \mathbb{E}_{(x,y) \sim \mathcal{D}} \sup_{\delta \in \BB_{\epsilon}} \left[ \frac{s(x,y)^\top \delta}{\|s(x,y)\|_2 \|\delta\|_2} \right] . 
\end{align*}



\paragraph{Perturbation sets with different degrees of alignment} 
Throughout this section, we intervene on the alignment of the perturbation set by restricting the threat model to \(k\) random orthogonal directions. 
More formally, we randomly sample \(k\) orthogonal unit vectors \(\delta_j\) and define the nested threat model as 
\begin{align}
\label{eq:sig_images}
	\BB_{\epsilon,k}= \cup_{j=1}^k \left\{\delta_j \beta  \mid |\beta|  \leq \epsilon\right\},
\end{align}
with $\BB_{\epsilon,1}\subset \BB_{\epsilon, 2} \dots $. 
It is important to note that as \(k\) increases, there are more choices for the direction of the adversarial perturbation and hence $\alignment(\BB_{\epsilon,k})$ increases with $k$ even though the perturbation budget $\epsilon$ stays constant.
Furthermore, unlike $\ell_p$-ball perturbations, we can evaluate the empirical robust error exactly for this threat model with a line-search along the $k$ random orthogonal directions.




\begin{figure}[t!]
    \centering
    \begin{subfigure}{0.48\textwidth}
        \includegraphics[width=\linewidth]{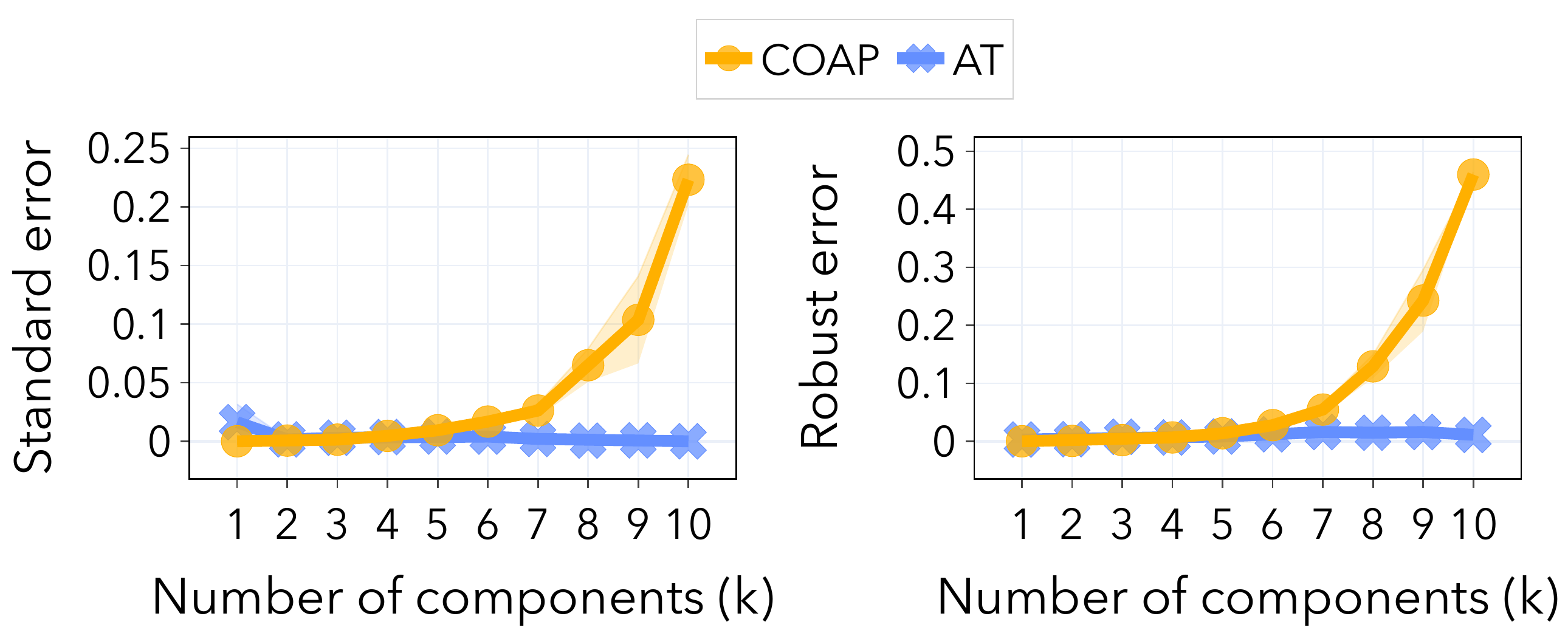}
        \caption{Concentric spheres}
        \label{fig:spheres_k}
    \end{subfigure}%
    \hfill
    \begin{subfigure}{0.48\textwidth}
        \includegraphics[width=\linewidth]{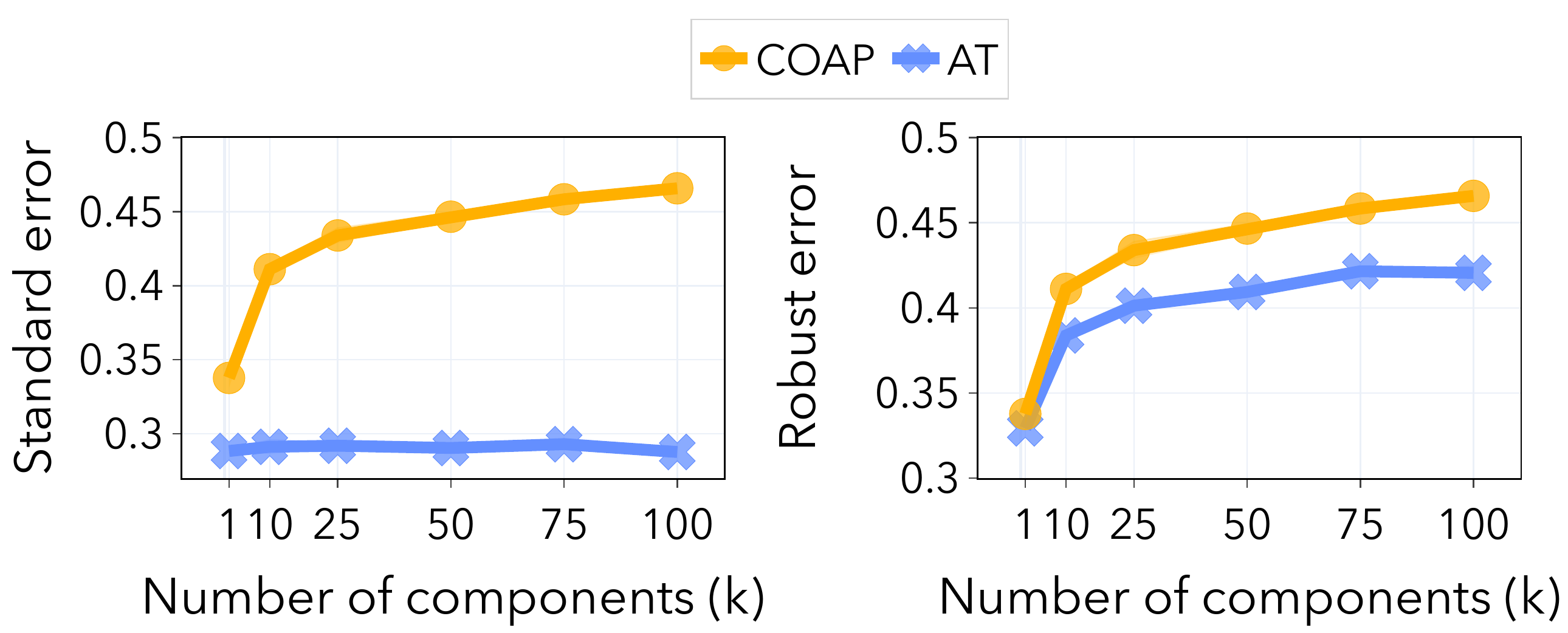}
        \caption{CIFAR-10}
        \label{fig:cifar_k}
    \end{subfigure}%
    \caption{Ablations for the signal aligned threat model $\BB_{\epsilon,k}$ defined in~\Cref{eq:sig_images}:      (a) We report mean and standard error over 5 runs for standard and robust error  on the concentric spheres dataset ($n=500, d=10, \gamma=20$, $\epsilon=5.0$). (b) We report mean and standard error over 3 runs for standard and robust error on the CIFAR-10 dataset ($\epsilon=2.5$). }
\end{figure}

\paragraph{Ablations} We now present a series of ablations on the number of components \(k\) to show that stronger alignment indeed correlates with a wider standard and robust error gap between \wong and \madry. 
In particular, in~\Cref{fig:spheres_k} we observe for the concentric spheres distribution that stronger alignment yields a gap of up to $25\%$ for standard error and $45\%$ for robust error. In~\Cref{fig:cifar_k}, we observe for CIFAR-10 a standard error gap of up to \(15\%\) and robust error gap of up to $5\%$, for the largest number of components.

\subsection{Factor (ii): Perturbation budget} 
\label{sec:fac2}

The second factor we investigate is the perturbation budget $\epsilon$ of the threat model. This quantity is directly related to the tightness of convex relaxations. 
In particular, \citet{wong_provable_2018} show that the \wong relaxation is very tight for small \(\epsilon\), while it becomes loose for larger values of \(\epsilon\). 
Although it is well-known that a larger perturbation budget results in worse certified training performance, the same also holds for adversarial training and it is unknown whether the error gap between the two increases.

\begin{figure}[tp!]
    \centering
    \begin{subfigure}{0.48\textwidth}
        \includegraphics[width=\linewidth]{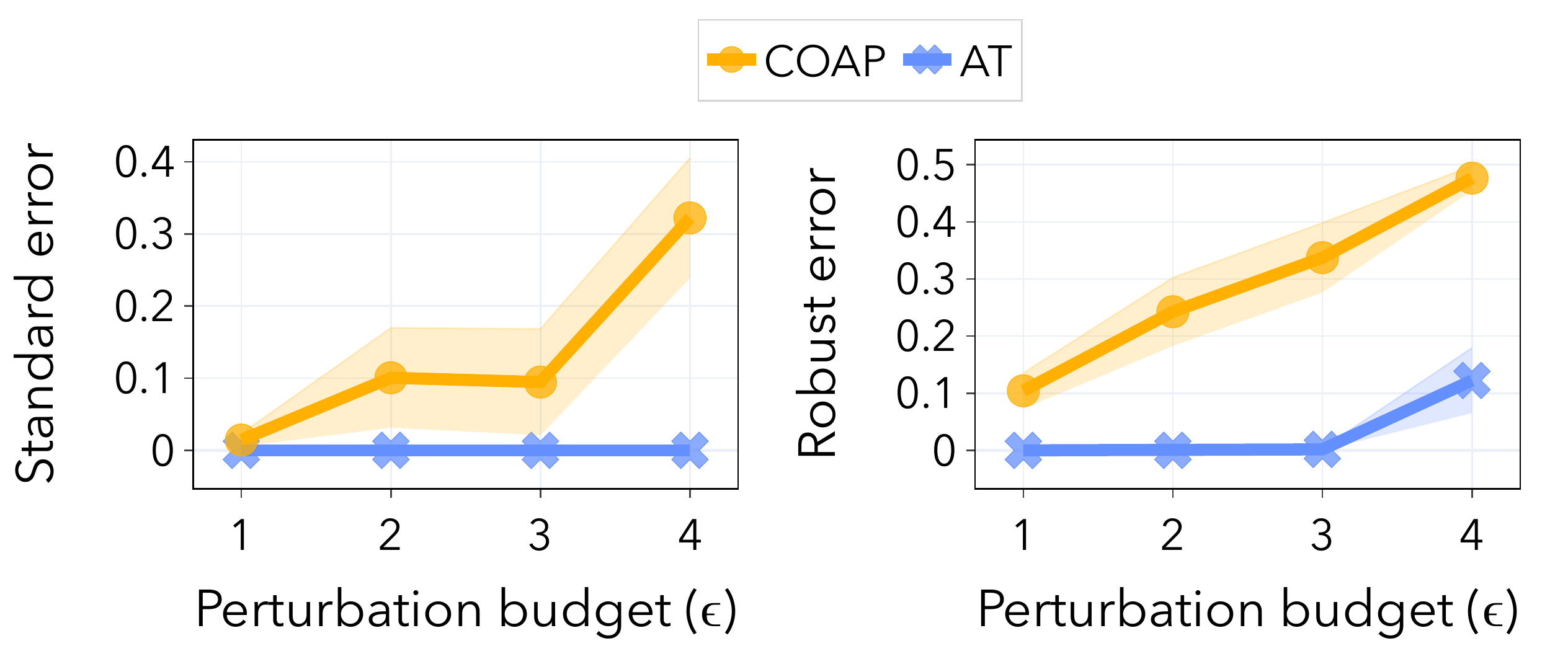}
        \caption{Concentric spheres}
        \label{fig:spheres_eps}
    \end{subfigure}%
    \hfill
    \begin{subfigure}{0.48\textwidth}
        \includegraphics[width=\linewidth]{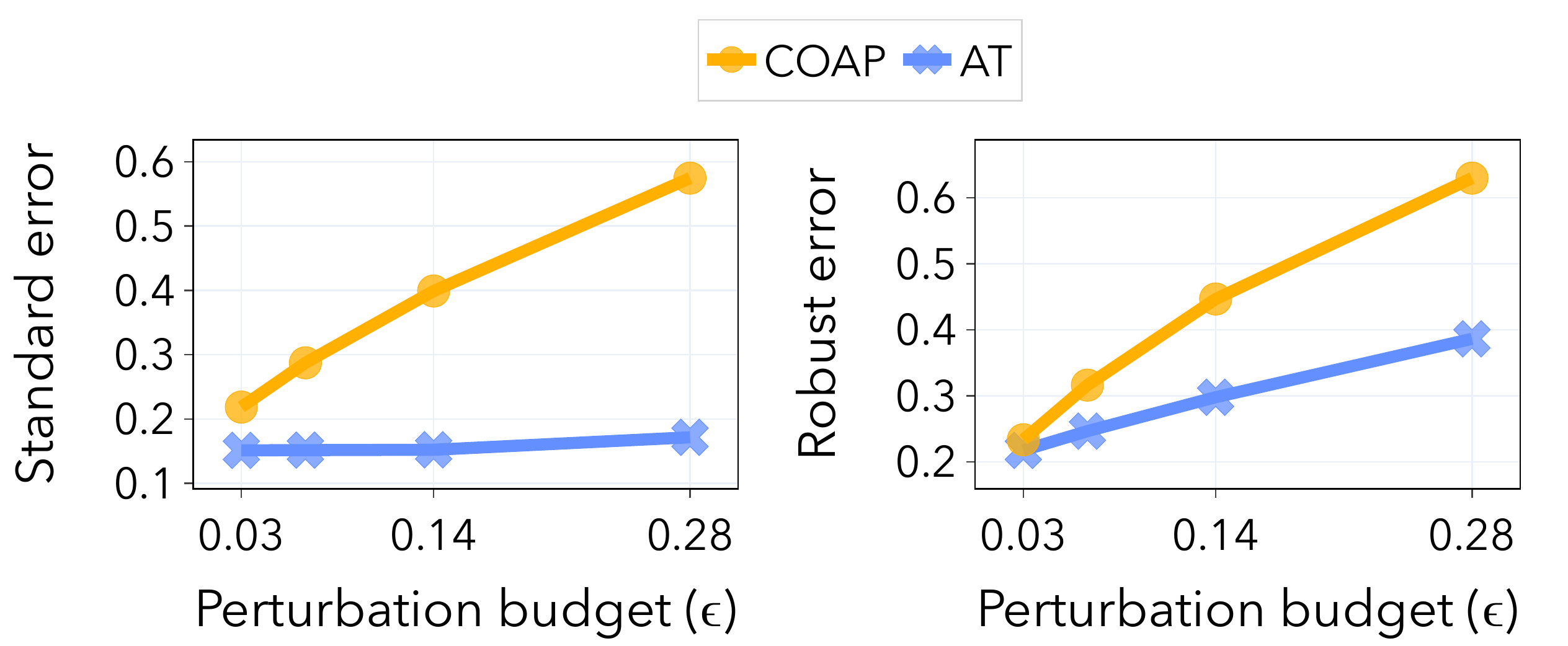}
        \caption{CIFAR-10}
        \label{fig:cifar_eps}
    \end{subfigure}%
    \caption{Ablations for the $\ell_2$-ball threat model: (a) We report mean and standard error over 5 runs for standard and robust error  on the concentric spheres dataset ($n=500,d=10, \gamma=20$). (b) Standard and robust error for \wong and \madry on the CIFAR-10 dataset.}
\end{figure}

\paragraph{Ablations} In~\Cref{fig:spheres_eps,fig:cifar_eps} 
we show that indeed, a larger perturbation budget correlates with a wider gap for both standard and robust error. 
 We observe that for both concentric spheres and CIFAR-10 the error gap monotonically increases as a function of the perturbation budget. Additionally, it is worth noting that this increase is not a consequence of a limited sample size. \wong training can indeed fail when the perturbation budget is significant, even given a large sample size. This is evident in~\Cref{fig:spheres_eps} for the case of concentric spheres, where the sample size is $n=500$ and the data dimensionality is $d=10$. \

\subsection{Factor (iii): Margin of the data distribution} 
\label{sec:fac3}
The third factor we investigate is the implicit margin of the data distribution, i.e. the minimum distance in feature space between any two data points of different classes.

 More formally, we define the margin \(\gamma\) for a dataset \(D = \{(x_i,y_i)\}_{i=1}^{n}\) as
\begin{align*}
	\gamma \defn \operatorname{min}_{i,j} \| x_i -x_j \|_2 \quad \text { s.t. } \quad i \neq j \;,\; y_i \neq y_j.
\end{align*}
In any robust classification task, this factor plays a pivotal role. When the implicit margin is small, the task becomes inherently more challenging, leading to potential failures in both adversarial and certified training. However, given that certified training essentially relies on an over-approximation of the perturbation set, it can be posited that it is particularly susceptible to failure for small margins. Below, we present as evidence an ablation study on the concentric spheres dataset. 

\begin{figure}[t!]
    \centering
    \begin{subfigure}{0.48\textwidth}
        \includegraphics[width=\linewidth]{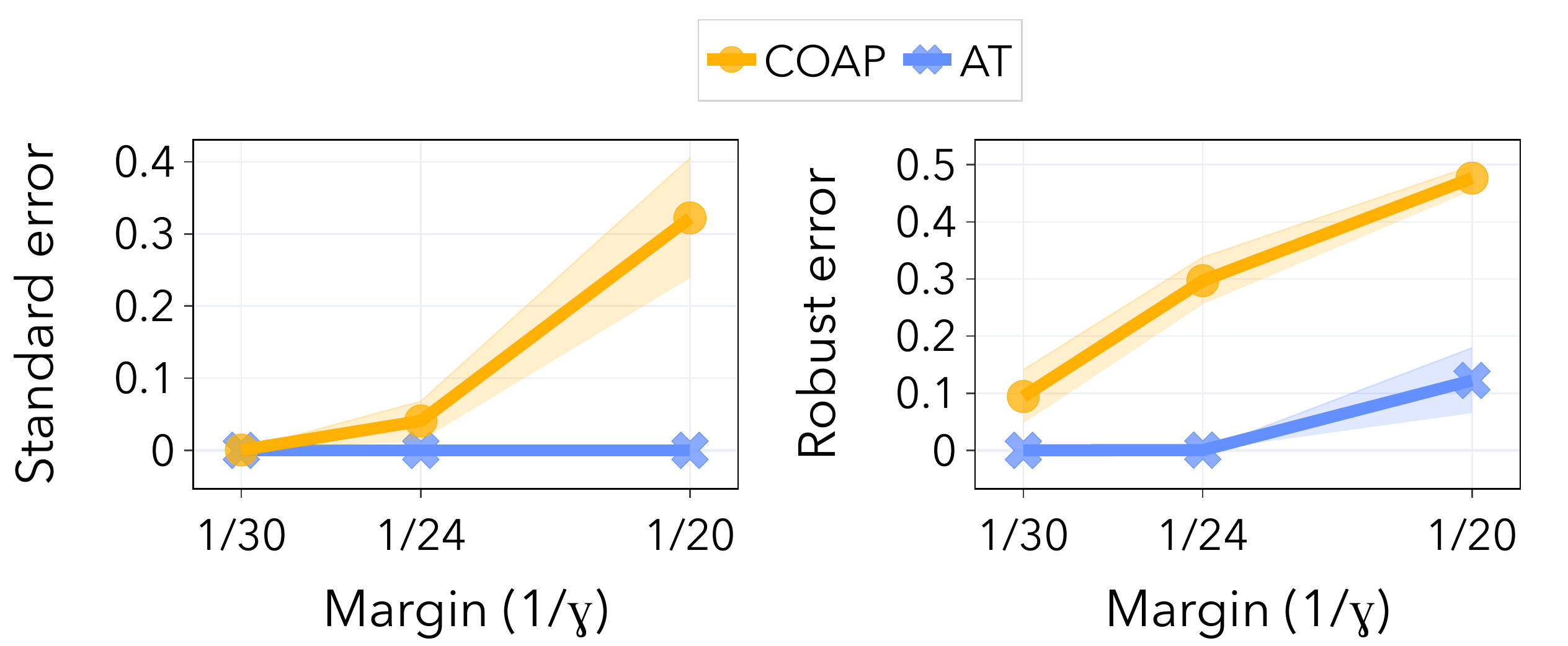}
        \caption{Concentric spheres}
        \label{fig:spheres_gamma}
    \end{subfigure}%
    \hfill
    \begin{subfigure}{0.35\textwidth}
        \includegraphics[scale=0.3]{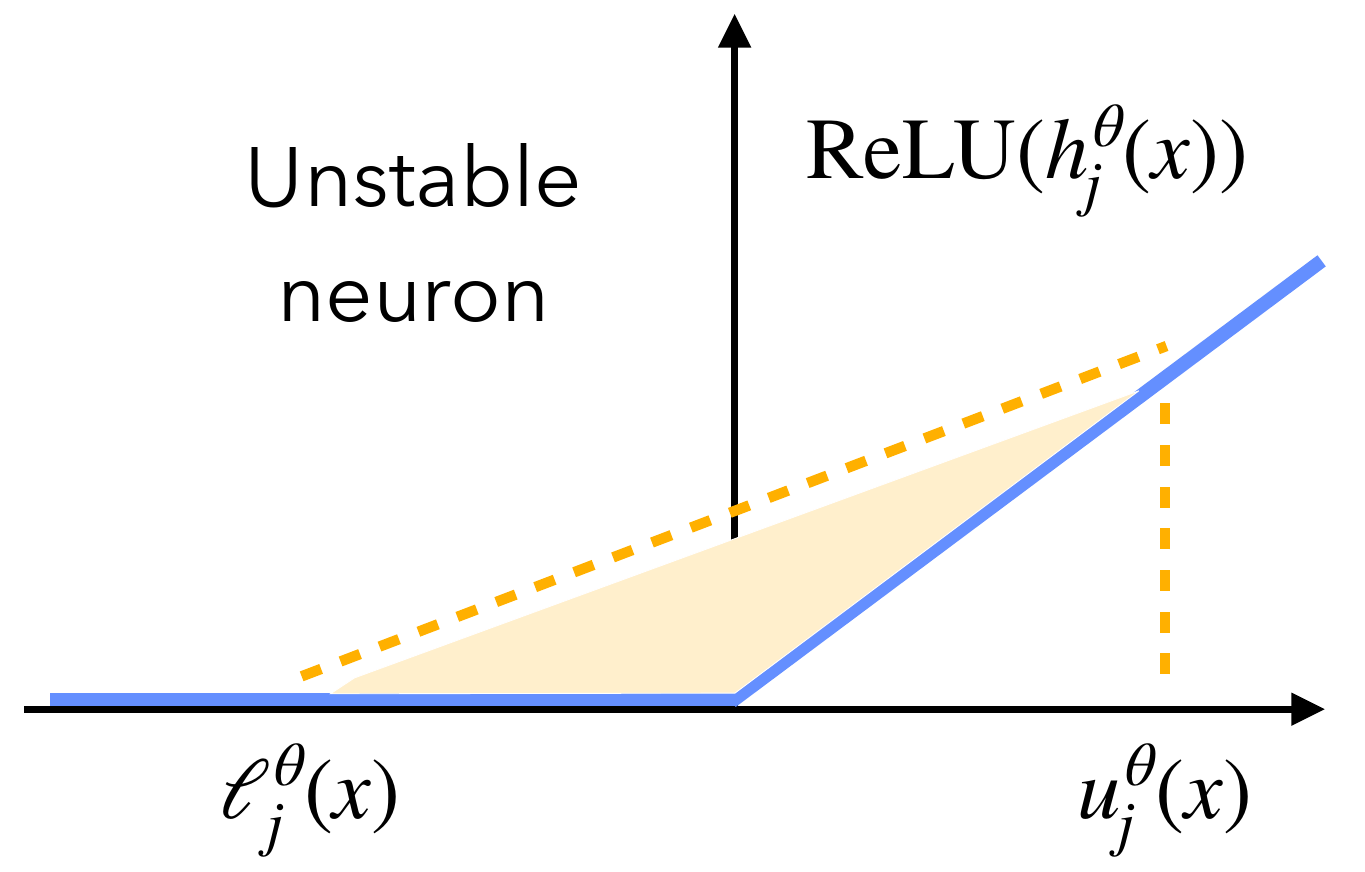}
        \caption{}
        \label{fig:relu_approximation}
    \end{subfigure}%
    \caption{(a) Ablations for the $\ell_2$-ball threat model:      We report mean and standard error over 5 runs for standard and robust error  on the concentric spheres dataset ($n=500, d=10$, $\epsilon=4.0$). (b) Conceptual illustration of the COAP convex relaxation for neurons in the unstable state.} 
\end{figure}

\paragraph{Ablations} We empirically show that a smaller margin correlates with a wider standard and robust error gap between \madry and \wong. 
As intervening on the implicit margin for image datasets such as CIFAR-10 is not feasible, we focus our ablations on
the concentric spheres dataset instead. In particular, in~\Cref{fig:spheres_gamma}, we observe that the standard and
robust error gap steadily increases as a function of the inverse margin, reaching up to \(30\%\) and \(35\%\), respectively.

\section{The role of unstable neurons}
\label{sec:neurons}
In this section, we attempt to explain how the previously identified factors influence the standard and robust error gap. Specifically, we discuss how they affect the error gap via their impact on the number of \emph{unstable neurons}, a quantity which plays a crucial role in the performance of certified training with convex relaxations.


\paragraph{Definition of unstable neurons} 
For a neural network $f_\theta$ with $m$ total number of neurons, we use
$h^\theta_{j}(x)$ to denote the pre-activation 
value for an input $x$ and a neuron $j$.
Similarly, we define lower and upper bounds for this value under perturbations
$\left[\plb_{j}^\theta(x),
\pub_{j}^\theta(x)\right]$, where $\plb_{j}^\theta(x) \leq
h_{j}^\theta(x+\delta) \leq \pub_{j}^\theta(x) $ 
for all allowed perturbations \(\delta \in \BB_\epsilon\).  Formally, we define \emph{inactive neurons} for an input $x$, as all neurons $j$
with non-positive pre-activation upper bounds $\pub_{j}^\theta(x) \leq 0$, i.e. they are always inactive regardless of input perturbations. Similarly, \emph{active neurons} have non-negative pre-activation lower bounds $\plb_{j}^\theta(x) \geq 0$, i.e. they are always active. In contrast, \emph{unstable} neurons have uncertain activation states given different input perturbations, i.e. $\plb_{j}^\theta(x) \leq 0 \leq \pub_{j}^\theta(x)$. 
Given a neural network $f_\theta$ and a dataset $D=\{(x_i,y_i)\}_{i=1}^n$, we define the number of unstable neurons as 
\begin{equation}
    \unstneur(f_\theta) = \frac{1}{mn}\sum_{i=1}^n\sum_{j=1}^\totneur \Indi \{ \plb_j^\theta(x_i) \leq 0 \leq \pub_{j}^\theta(x_i)\}.
\end{equation}

\paragraph{Unstable neurons and convex relaxations} A key property of certified training with convex relaxations is the tightness of the over-approximation compared to the original perturbation set.
Since convex relaxations are much looser for unstable neurons compared to active or inactive neurons (exemplarily illustrated for the COAP convex relaxation~\citep{wong_scaling_2018} in~\Cref{fig:relu_approximation}), the number of unstable neurons is a good indicator for quantifying the looseness of the over-approximation of the perturbation set. 


Further, existing works~\citep{lee2021towards,shi2021fast, muller2023certified} suggest that the number of unstable neurons directly affects the performance of certified training. Intuitively, the looser the over-approximation during training, the greater the susceptibility to noise becomes. In particular, noise can be introduced during training  when the over-approximated perturbation set extends over the (robust Bayes optimal) decision boundary, essentially causing over-regularization. Hence, it is natural to expect that 
an increased number of unstable neurons during training
might cause a larger  error gap between certified and adversarial training. However, it is impossible to verify this hypothesis by only intervening on the number of unstable neurons in a neural network  while keeping all other factors fixed.

For the above reasons, we study how the factors identified in~\Cref{sec:factors} affect the total number of unstable neurons during training, i.e. $\sum_{t=1}^T \unstneur(f_{\theta^t})$ where $f_{\theta^t}$ is the neural network at epoch $t$ and $T$ is the total number of training epochs.   In particular, we first provide empirical evidence that the three factors correlate with a higher number of unstable neurons. Then, we provide an intuitive explanation with illustrations on a simple realisation of the concentric spheres distribution.


\begin{figure}[t]
    \centering
\begin{subfigure}{\textwidth}

\includegraphics[width=\linewidth]{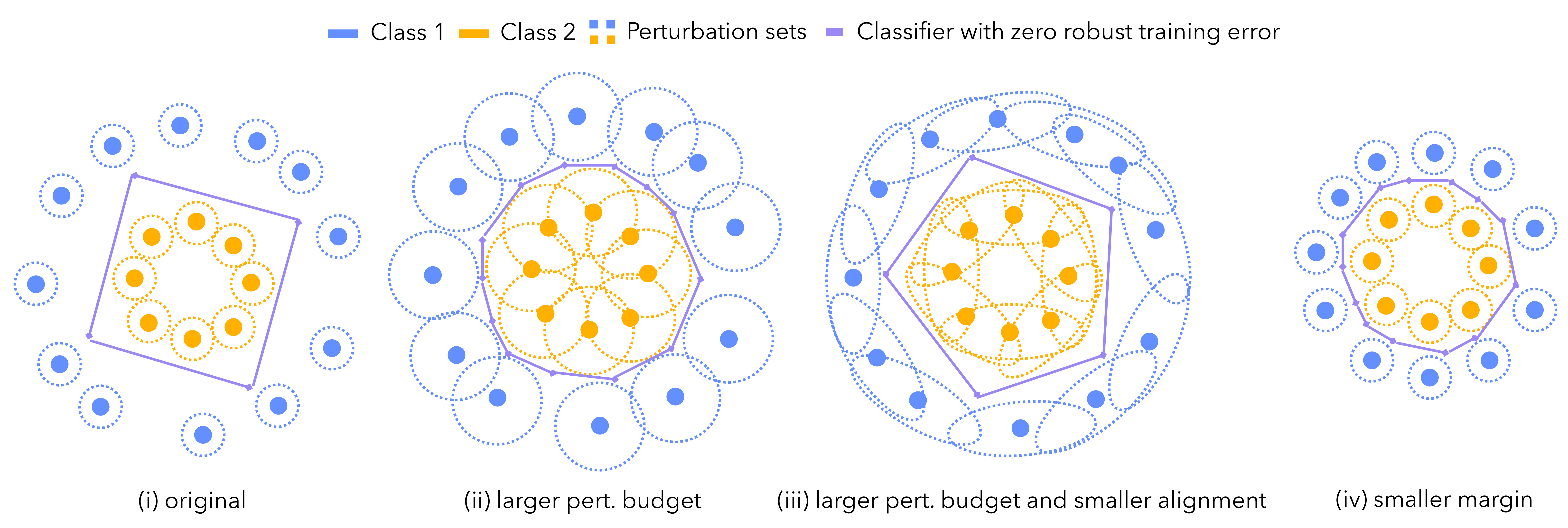}
        \caption{}
\label{fig:spheres_intuition}
\end{subfigure}
    \centering
      \begin{subfigure}{0.2\textwidth}
        \includegraphics[width=\linewidth]{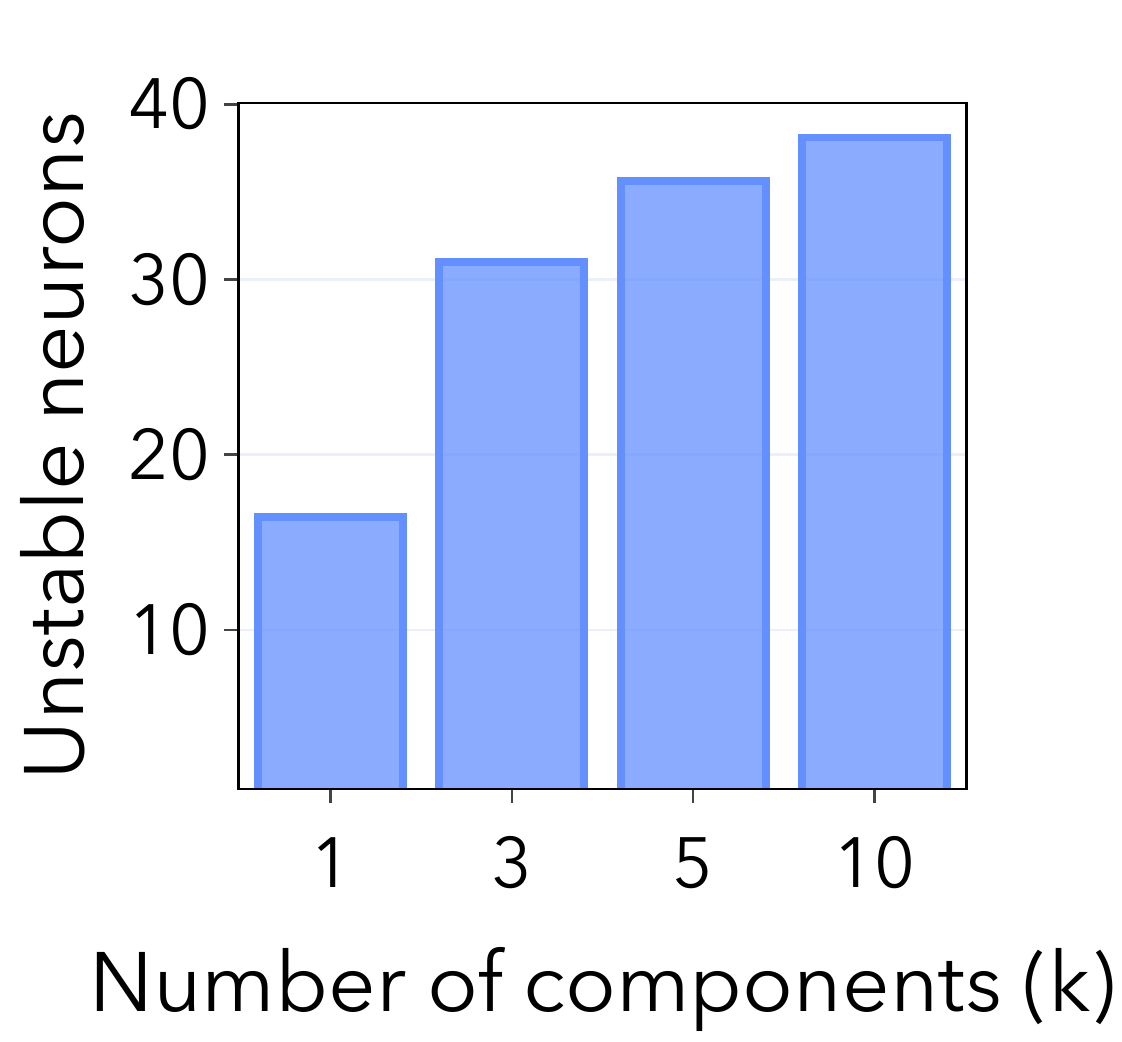}
        \caption{Conc. spheres}
        \label{fig:spheres_k_vs_un}
    \end{subfigure}%
    \hfill 
    \begin{subfigure}{0.2\textwidth}
        \includegraphics[width=\linewidth]{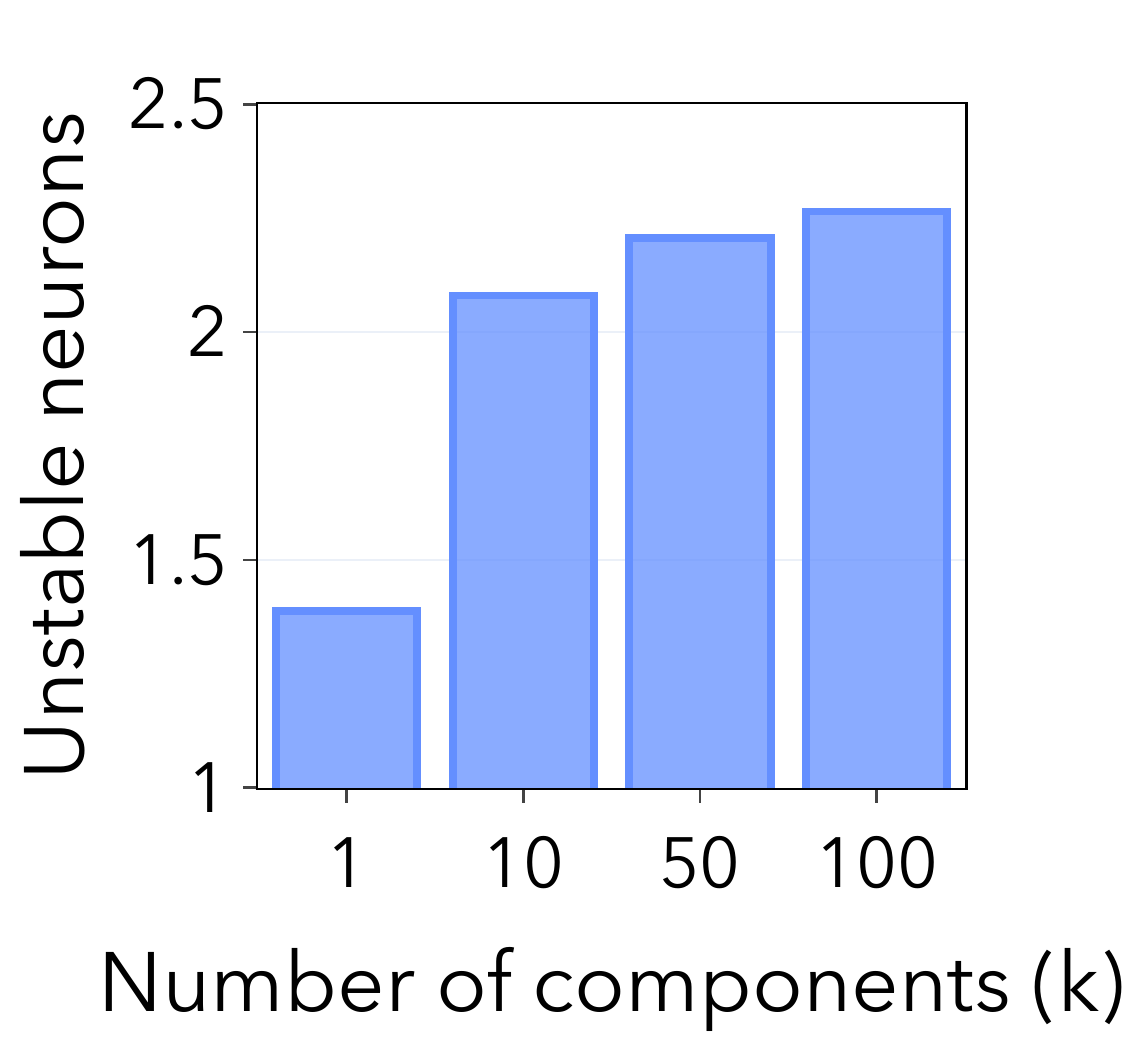}
        \caption{CIFAR-10}
        \label{fig:cifar_k_vs_un}
    \end{subfigure}%
    \hfill
        \begin{subfigure}{0.2\textwidth}
        \includegraphics[width=\linewidth]{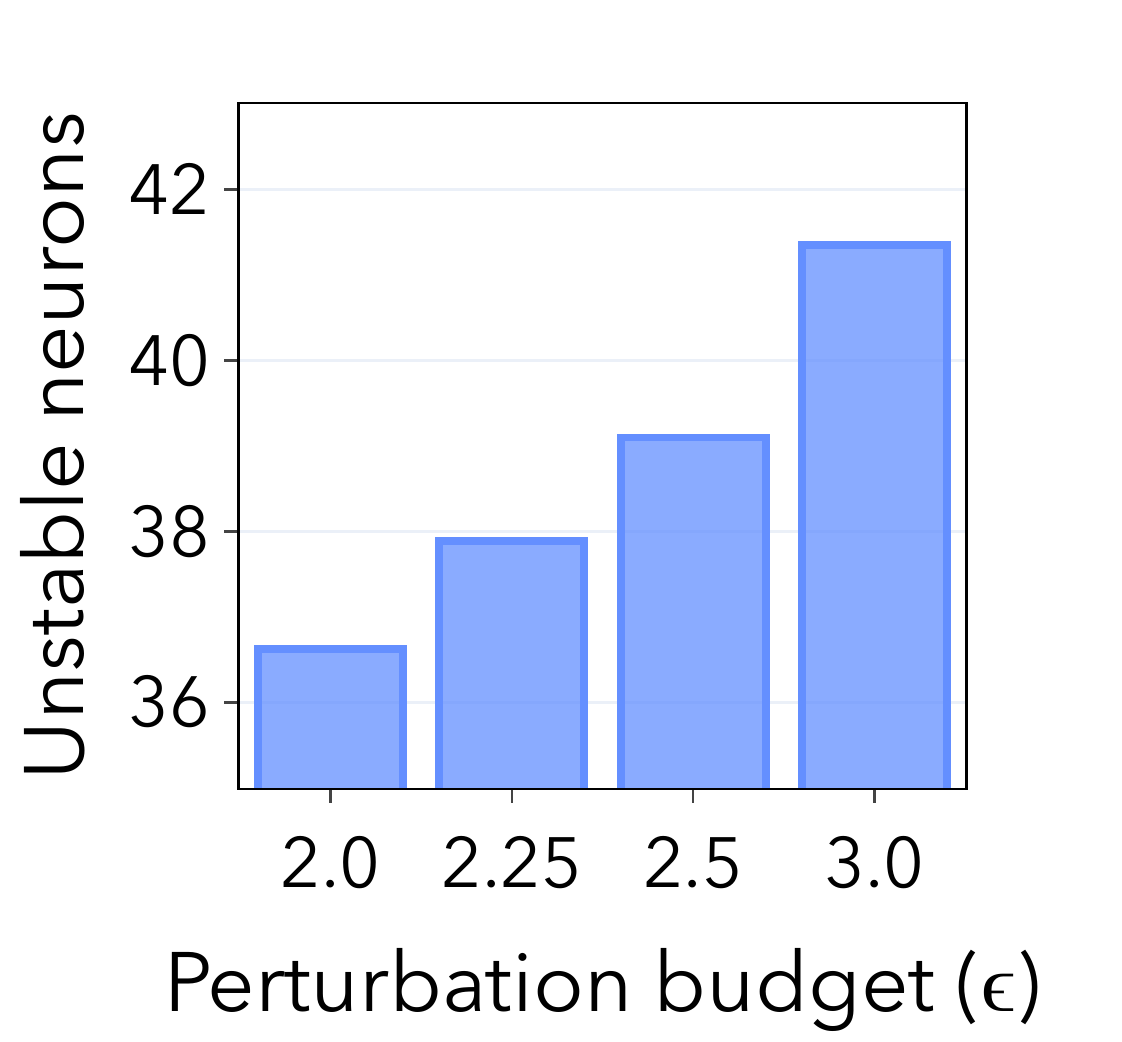}
        \caption{Conc. spheres}
        \label{fig:spheres_eps_vs_un}
    \end{subfigure}%
        \hfill
     \begin{subfigure}{0.2\textwidth}
        \includegraphics[width=\linewidth]{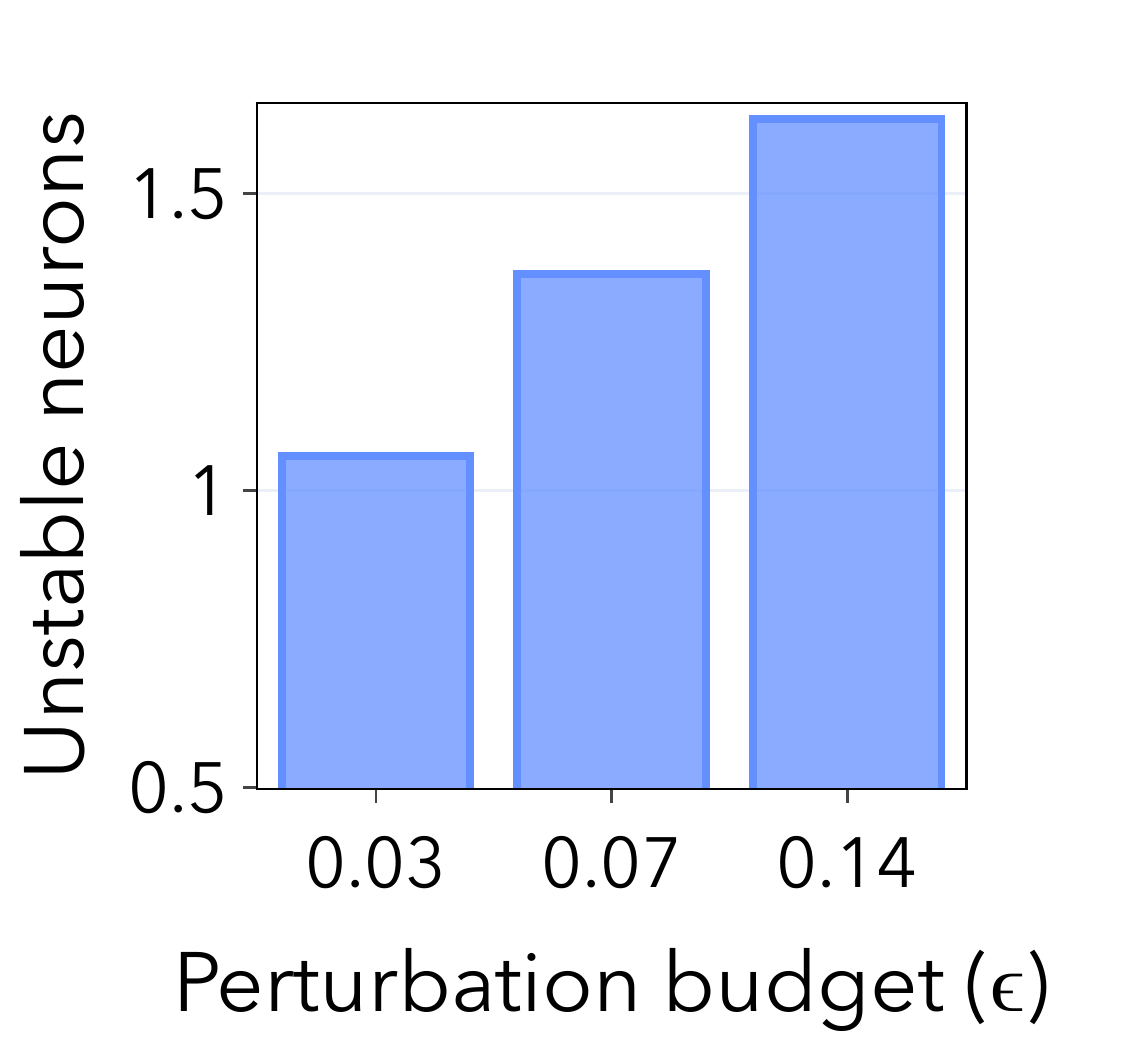}
        \caption{CIFAR-10}
        \label{fig:cifar_eps_vs_un}
    \end{subfigure}%
    \hfill
    \begin{subfigure}{0.2\textwidth}
        \includegraphics[width=\linewidth]{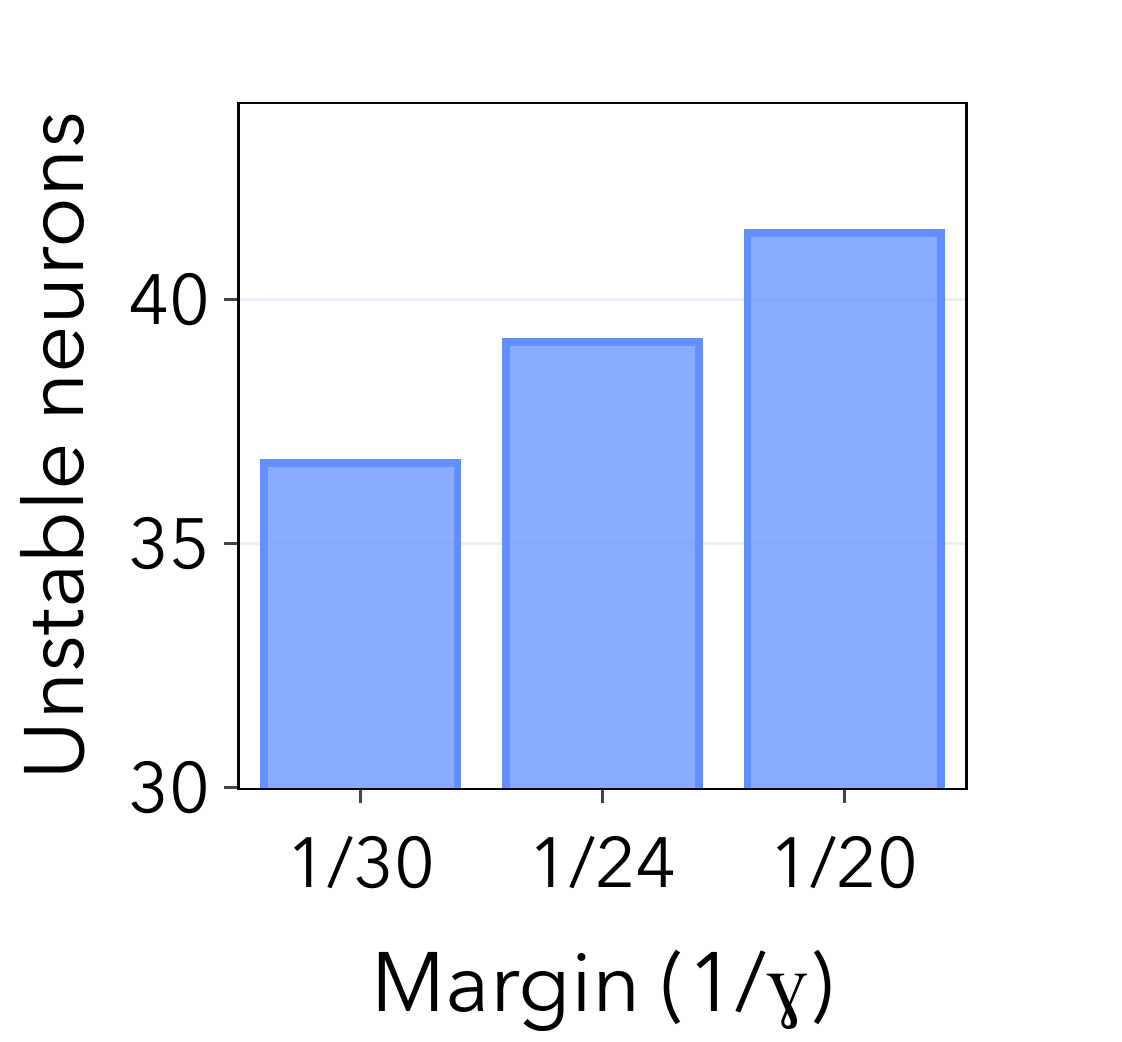}
        \caption{Conc. spheres}
        \label{fig:spheres_gamma_vs_un}
    \end{subfigure}%

    \caption{(a) Conceptual illustration of how the three factors act on the number of unstable neurons.  (b -- f) Ablations for the signal aligned and the $\ell_2$-ball threat models: (b,d,f) Concentric spheres dataset ($\gamma=20, n=500, d=10,  \epsilon_2=5.0, \epsilon_{\text{signal}}=3.0$). (c,e) CIFAR-10 dataset ($\epsilon_{\text{signal}}=2.5$).}
\end{figure}
\paragraph{How the factors affect unstable neurons}
In~\Cref{fig:spheres_k_vs_un,fig:cifar_k_vs_un,fig:spheres_eps_vs_un,fig:cifar_eps_vs_un,fig:spheres_gamma_vs_un}, we can observe how for both the CIFAR-10 and the concentric spheres datasets, the number of unstable neurons increases with larger alignment, larger perturbation budget and  smaller margin.   


We interpret these findings as follows: both smaller margin and larger perturbation budget increase the model complexity of the decision boundary with good robust training accuracy (similar to an argument made by \citet{nakkiran19}). In particular, they require a larger number of piecewise linear regions, as also argued in~\citet{shah2020pitfalls}.
This intuition is illustrated in~\Cref{fig:spheres_intuition}: for a fixed perturbation set, both smaller margin (iv) and  larger perturbation budget (ii)  require a larger number of piecewise linear functions to approximate the decision boundary. 
The increased number of linear regions, together with training points close to the boundary, then results in an increase in the number of unstable neurons. 
Further, for less aligned perturbation sets with the same margin and perturbation budget, i.e. contrasting (iii) and (ii), the learning task is simplified, and fewer piecewise regions are required. 
As a result,  fewer neurons become unstable.



\section{Related work}
\label{sec:related}

\paragraph{Limitations of certified training with convex relaxations} 
Certified training with convex relaxations hinges on over-approximating the potential output range of each neuron for the perturbed versions of any input point.
While this over-approximation allows for tractable computation of an upper bound on the robust error, it also introduces an inherent looseness that impacts both robust training and evaluation. For example, \citet{salman_convex_2019} investigate the tightness of convex relaxations for verification purposes, i.e. for certifying the robustness of already trained models, and show that even the best existing convex relaxation provides only very loose upper bounds on the robust error. 

The effect of the tightness of convex relaxations on certified training has recently been studied in the context of the so-called paradox of certified training~\citep{jovanovic_paradox_2021}, i.e. when training with tighter relaxations leads to worse certified robustness. 
\citet{lee2021towards} show empirically that tighter convex relaxations affect the smoothness of the loss function, which in turn impacts the performance of certified training.  
Instead of comparing convex relaxations with different tightness, we discuss how tightness affects the performance of certified training  \emph{in the context} of varying other factors related to the threat model and the data distribution.

\paragraph{Limitations of randomised smoothing} As an alternative to convex relaxations, randomised smoothing gives robustness guarantees with a certain probability~\citep{cohen_certified_2019,lecuyer_certified_2019,li_certified_2019}. 
Despite its popularity, randomised
smoothing also suffers from several limitations beyond an increased
computational cost. 
For example, several works have exposed an accuracy-robustness tradeoff ~\citep{blum_random_2020,
kumar_curse_2020}. Further, randomised smoothing can significantly hurt the disparity in class-wise accuracy~\citep{mohapatra_hidden_2021} and is extremely vulnerable to low-frequency corruptions of the test data~\citep{avidan_spectral_2022}. In general, compared to certified training with convex relaxations, the drawbacks of randomised smoothing are much better understood, and efforts are being made towards developing new defences to bridge the gap with adversarial training~\citep{nandy_towards_2022}. Therefore, we focus our attention in this paper on certified training with convex relaxations. 
\paragraph{Limitations of adversarial training} Not only certified training comes at the price of reduced test performance - the same has been reported for adversarial training. For example, it has been well-studied that adversarial training often results in a trade-off between robust and standard accuracy; that is, the standard accuracy of adversarially trained models often decreases even though the robust accuracy increases (see, e.g. ~\citep{tsipras_robustness_2019,
zhang_theoretically_2019,raghunathan_understanding_2020}). 
Attempts to provide an explanation for consistent perturbations, i.e. perturbations that do not change the true label, have so far focused on the small sample size regime. For example,  \citet{raghunathan_understanding_2020} prove that for small sample sizes, adversarial training can increase standard error even in the absence of noise. Further, most related to our work, \citet{clarysse_why_2022} recently prove that adversarial training with perturbations aligned with the signal direction can even increase the robust error. 
In contrast to this line of work, our experiments focus on the large-sample regime and compare certified with adversarial training.

\section{Conclusion}
In this paper, we show that
 	certified training with convex relaxations suffers from worse standard error and robust error than adversarial training. Further, we are the first to provide a systematic comparison of these two robust training paradigms across multiple datasets and threat models. In doing so, we explore three important factors which are correlated with a wider standard and robust error gap between certified and adversarial training. We believe that shedding light on this error gap will not only provide us with a clearer picture of the trade-offs observed in practice but also lead to better approaches for certified robustness.
\section*{Acknowledgements}
PDB was supported by the  Hasler Foundation grant number 21050 and AS acknowledges partial support from the ETH AI Center postdoctoral fellowship. 

\bibliography{refs}
\bibliographystyle{abbrvnat}

\clearpage
\appendix

\section{Experimental Details}
\label{apx:exp_details}
\subsection{Image experiments with \texorpdfstring{$\ell_2$-ball}{l2-ball} and \texorpdfstring{$\ell_\infty$-ball}{linf-ball} threat models}
\label{apx:image_experiments}
Below we provide complete experimental details to reproduce: \Cref{table:image_experiments,table:mnist,table:cifar,fig:cifar_eps,fig:cifar_eps_vs_un}.

\paragraph{Model architectures} For MNIST, we train the CNN architecture with four convolutional layer and two fully connected layers of 512 units introduced in~\citet{wong_scaling_2018}. We report the architectural details in~\Cref{table:mnist_cifar_arch}. For CIFAR-10, we  train the residual network (ResNet) with the same structure used in  \citet{wong_scaling_2018};  we use 1 residual block with 16, 16, 32, and 64 filters. For Tiny ImageNet, we train a  WideResNet  as in~\citet{xu_automatic_2020}, using three wide basic blocks with a wide factor of 10.

\begin{table}[h]
\begin{center}
\begin{small}
\begin{tabular}{lccc}
\toprule
 CNN  \\
\midrule
   CONV $32 \; 3 \times 3+1$ \\
 CONV $32 \; 4 \times 4+2$  \\
CONV $64 \; 3 \times 3+1$  \\
 CONV $64 \;4 \times 4+2$  \\
FC 512 \\
 FC 512 \\
 \bottomrule
\end{tabular}
\end{small}
\end{center}
\caption{MNIST model architecture. All layers are followed by $\relu{\cdot}$ activations. The last fully connected layer is omitted. "CONV $k$ $w \times h +s$" corresponds to a 2D convolutional layer with k filters of size $w \times h$  using a stride of $s$ in both dimensions. "FC $n$" is a fully connected layer with $n$ outputs.}
\label{table:mnist_cifar_arch}
\end{table}

\paragraph{Dataset preprocessing}

For MNIST, we use full $28\times28$ images without any augmentations and normalisation. For CIFAR-10, we use random horizontal flips and random crops as data augmentation, and normalise images according to per-channel statistics. For Tiny ImageNet, we use random crops of 56 \(\times\) 56 and random flips during training. During testing, we use a central 56 \(\times\) 56 crop. We also normalise images according to per-channel statistics.

\paragraph{Robust evaluation}
We consider $\ell_2$-ball perturbations. We evaluate the robust error using the most expensive version of AutoAttack (AA+)~\citep{croce_reliable_2020}. Specifically, we include the following attacks:
untargeted APGD-CE (5 restarts), untargeted APGD-DLR (5 restarts), untargeted APGD-DLR (5 restarts), Square Attack (5000 queries), targeted APGD-DLR (9 target classes) and targeted FAB (9 target classes).

 \paragraph{AT training details} For MNIST, we train 100 epochs using   Adam optimiser~\citep{kingma_adam_2015} with a learning rate of $0.001$, momentum of $0.9$ and a  batch size of 128; we reduce the learning rate by a factor 0.1 at epochs 40 and 80. For CIFAR-10 with ResNet, we train 150 epochs using SGD with a learning rate of $0.05$ and a batch size of 128;  we reduce the learning rate by a factor of 0.1 at epochs 80 and 120. For Tiny ImageNet and CIFAR-10 with Wide-Resnet we train 200 epochs using SGD with a learning rate of 0.1 and a batch size of 512; we reduce the learning rate by a factor of 0.1 at epochs 100 and 150. For the inner optimisation of all models and datasets, adversarial examples are generated with 10 iterations of  Auto-PGD~\citep{croce_reliable_2020}. 

 \paragraph{COAP training details}  We follow the settings proposed by the authors and report them here. For  MNIST, we use the Adam optimiser~\citep{kingma_adam_2015} with a learning rate of 0.001 and a batch size of 50. We schedule $\epsilon$ starting from 0.01 to the desired value over the first 20 epochs, after which we decay the learning rate by a factor of 0.5 every 10 epochs for a total of 60 epochs. For  CIFAR-10, we use the SGD optimiser with a learning rate of 0.05 and a batch size of 50. We schedule $\epsilon$ starting from 0.001 to the desired value over the first 20 epochs, after which we decay the learning rate by a factor of 0.5 every 10 epochs for a total of 60 epochs.  For all datasets and models, we use random projection of 50 dimensions. For all experiments, we use the implementation provided in~\citet{wong_scaling_2018}.

\paragraph{CROWN-IBP and IBP training details} We follow the settings proposed by the authors and report them here. For MNIST, we train 200 epochs with a batch size of 256. We use Adam optimiser~\citep{kingma_adam_2015} and set the learning rate to $5 \times 10^{-4}$. We warm up with 10 epochs of regular training 
 and gradually ramp up $\epsilon_{\text{train}}$ from 0 to $\epsilon$ in 50 epochs. We reduce the learning rate by a factor of 0.1 at epochs 130 and 190. 
 For CIFAR-10, we train 2000 epochs with a batch size of 256 and a learning rate of $5 \times 10^{-4}$. We warm up for 100 epochs and ramp-up $\epsilon$ for 800 epochs.
 The learning rate is reduced by a factor of $0.1$ at epochs 1400 and 1700. For Tiny ImageNet, we train 600 epochs with batch size 128. The first 100 epochs are clean training. Then we gradually increase $\epsilon_{\text{train}}$ with a schedule length of 400. For all datasets, a hyper-parameter $\beta$ to balance LiRPA bounds and IBP bounds for the output layer is gradually decreased from 1 to 0 (1 for only using LiRPA bounds and 0 for only using IBP bounds), with the same schedule of $\epsilon$. For all experiments,  we use the implementation provided in the auto LiRPA library~\citep{xu_automatic_2020}.

 \paragraph{FAST-IBP training details} We follow the settings proposed by the authors and report them here. Further, we modify the architecture to add batch normalisation at each layer, as suggested by the authors.  
 Models are trained with Adam optimiser~\citep{kingma_adam_2015} with an initial learning rate of $5 \times 10^{-4}$, and there are two milestones where the learning rate decays by 0.2. We determine the milestones for learning rate decay according to the training schedule and the total number of epochs, as shown in~\Cref{table:fastibp_details}. The gradient clipping threshold is set to 10.0. We train the models using a batch size of 256 on MNIST and 128 on CIFAR-10 and TinyImageNet. The tolerance value $\tau$ in our warmup regularization is fixed to 0.5.

During the warmup stage, after training with $\epsilon=0$ for a number of epochs, the perturbation radius $\epsilon$ is gradually increased from 0 until the target perturbation radius $\epsilon_{\text {target }}$, during the $0<\epsilon<\epsilon_{\text {target }}$ phase. Specifically, during the first $25 \%$ epochs of the $\epsilon$ increasing stage, $\epsilon$ is increased exponentially, and after that $\epsilon$ is increased linearly. In this way, $\epsilon$ remains relatively small and increases relatively slowly during the beginning to stabilize training. We use the SmooothedScheduler in the autoLiRPA library as the scheduler for $\epsilon$, similarly adopted by~\citet{xu_automatic_2020}. 

\begin{table}[h]
\centering
\begin{small}
\begin{tabular}{lcccc}
\toprule
Dataset & Total epochs & Decay-1 & Decay-2 \\
\midrule
\multirow{1}{*}{MNIST} 
& 70 & 50 & 60 \\
\midrule
\multirow{1}{*}{CIFAR-10} & 160 & 120 & 140 \\
\midrule
TinyImageNet & 80 & 60 & 70 \\
\bottomrule
\end{tabular}
\vspace{0.1in}
\caption{Milestones for learning rate decay when the different total number of epochs are used. "Decay-1" and "Decay-2" denote the two milestones, respectively, when the learning rate decays by a factor of 0.2.}
\label{table:fastibp_details}
\end{small}
\end{table}

\subsection{Image experiments with signal aligned threat model}
Below we provide complete experimental details to reproduce:
\Cref{fig:cifar_k,fig:cifar_k_vs_un,fig:mnist_k_vs_error}.
First, we present our extension of \wong to the threat model introduced in~\Cref{eq:sig_images}. Rather than deriving the dual problem as in~\citet{wong_provable_2018}, we  consider the conjugate function view introduced in~\citet{wong_scaling_2018}. In particular, we only have to modify the dual of the input layer to the network. Below we derive the  conjugate bound for the signal-aligned threat model:
\begin{align*}
	\sup_{\delta \in \BB_{\epsilon,k}}  \nu_1^\top (x+\delta)  &= \sup_{k,\beta}    \nu_1^\top(x+s_k \beta) \\&= \nu_1^\top x + \epsilon\max_{k} |\nu_1^\top s_k|
\end{align*}
For all experiments, we use the convolutional neural network architecture described in~\Cref{table:mnist_cifar_arch}. Note that it is not possible to scale to ResNet with the threat model in~\Cref{eq:sig_images}, as the random projections trick derived in~\citet{wong_scaling_2018} is tailored to \(\ell_\infty\)-ball and \(\ell_2\)-ball threat models.

 \paragraph{AT training details} For both MNIST and CIFAR-10, we train 20 epochs using Adam optimiser~\citep{kingma_adam_2015} with a learning rate of $0.001$, momentum of $0.9$ and a  batch size of 64; we reduce the learning rate by a factor 0.1 at epochs 10. For the inner optimisation of all models and datasets, we solve the exact problem as it is computationally efficient to line-search the maximal perturbation. 

 \paragraph{COAP training details} For both MNIST and CIFAR-10, we use the Adam optimiser~\citep{kingma_adam_2015} with a learning rate of 0.001 and a batch size of 64. We schedule $\epsilon$ starting from 0.01 to the desired value over the first 3 epochs, after which we decay the learning rate by a factor of 0.5 every 10 epochs. For all datasets and models, we do not use random projections. For all experiments, we use the implementation provided in~\citet{wong_scaling_2018}.

 \subsection{Synthetic experiments with signal-aligned, \texorpdfstring{$\ell_2$}{l2}-ball and \texorpdfstring{$\ell_\infty$}{linf}-ball threat models}
\label{apx:synthetic_exp_l2}

Below we provide complete experimental details to reproduce \Cref{fig:spheres_k,fig:spheres_eps,fig:spheres_gamma,fig:spheres_gamma_vs_un,fig:spheres_eps_vs_un,fig:spheres_k_vs_un}.

\paragraph{Data generation} For the spheres dataset, we generate a random $x \in \RR^\di$ where $\norm{x}_2$ is either $R_0$ or $R_1$  , with equal probability assigned to each norm. 
We associate with each $x$ a label $y$ such that $y = -1$ if $\norm{x}_2= R_0$ and $y = 1$ if $\norm{x}_2 = R_1$. We can  sample uniformly from this distribution by sampling $z \sim \gauss\left(0,I_\di\right)$ and then setting $x = \frac{z}{\norm{z}_2}R_0$ or $x =\frac{z}{\norm{z}_2} R_1$. For all experiments with the concentric spheres distribution, we set $\di=10, \nsamples=500, \ntest=10^4$.

\paragraph{Model and hyper-parameters} For all the experiments, we use an MLP architecture with $\weight=100$ neurons and one hidden layer and $\relu{\cdot}$  activation functions. We use PyTorch SGD optimiser with a momentum of $0.95$ and train the network for 150 epochs. 
We sweep over the learning rate $\eta \in \{0.1,0.01,0.001\}$, and for each  perturbation budget, we choose the one that minimises robust error on the test set. 

\paragraph{Robust evaluation}  We  evaluate robust error at test-time using Auto-PGD~\citep{croce_reliable_2020} with 100 iterations and 5 random restarts. We use both the cross-entropy and the difference of logit loss to prevent gradient masking. We use the implementation provided in AutoAttack~\citep{croce_reliable_2020} with minor adjustments to allow for non-image inputs. 

\paragraph{Training paradigms}
For standard training~(ST), we train the network to minimise the cross-entropy loss. For adversarial training~(\madry)~\citep{madry_towards_2018, goodfellow_explaining_2015}, we train the network to minimise the robust cross-entropy loss. At each epoch, we search for adversarial examples using Auto-PGD~\citep{croce_reliable_2020}	 with a budget of 10 steps and 1 random restart. Then, we  update the weights using a gradient with respect to the adversarial examples. For convex outer adversarial polytope~(\wong)~\citep{wong_provable_2018, wong_scaling_2018}, we train the network to minimise the upper bound on the robust error. Our implementation is based on the code released by the authors.
 \clearpage
 
\section{Experimental comparison of certified and adversarial training}
\label{apx:complete_comparison}
This section provides a more complete evaluation of certified and adversarial training, including a wide range of perturbation budgets.

\paragraph{MNIST} Similar to \Cref{sec:images_evaluation}, we observe that for the $\ell_\infty$-ball threat model, FAST-IBP achieves the best robust error across all perturbation budgets, while \madry delivers the best standard error. On the other hand, for the $\ell_2$-ball threat model, \madry achieves both the best robust and standard error across all perturbation budgets.

\begin{table}[h]
\begin{center}
\begin{adjustbox}{width=0.7\textwidth}
\begin{small}
\begin{sc}
\begin{tabular}{lcccc}
\toprule
  Dataset & Perturbation budget& Method & Robust error & Standard error   \\
\midrule
    & & \madry & $0.021$ & $\mathbf{0.008}$ \\
     & & \wong & $0.026$ & $0.010$ \\
   MNIST & $\epsilon_\infty=0.1$ & CROWN-IBP & $0.024$ &  $0.010$\\
     & & IBP & $0.026$ & $0.010$ \\
     & & FAST-IBP & $\mathbf{0.017}$ & $0.009$ \\
     \midrule
&  & \madry & $0.045$ & $\mathbf{0.008}$ \\
& & \wong &  $0.066$ & $0.026$ \\
MNIST & $\epsilon_\infty=0.2$ & CROWN-IBP & $0.050$ & $0.014$ \\
 & & IBP & $0.060$  & $0.014$ \\
 & & FAST-IBP & $\mathbf{0.032}$ & $0.010$\\
\midrule

& & \madry & $0.095$ & $\mathbf{0.008}$ \\
 & & \wong & $0.224$ & $0.110$ \\
MNIST & $\epsilon_\infty=0.3$ & CROWN-IBP & $0.075$ & $0.017$ \\
&  & IBP & $0.091$ & $0.021$ \\
 & & FAST-IBP & $\mathbf{0.058}$ & $0.015$ \\
\midrule

 & & \madry & $0.321$ & $\mathbf{0.008}$ \\
 & & \wong & $0.806$ & $0.737$ \\
MNIST & $\epsilon_\infty=0.4$ & CROWN-IBP & $0.120$ & $0.025$ \\
 & & IBP & $0.147$ & $0.033$ \\
 & & FAST-IBP & $\mathbf{0.104}$ & $0.024$ \\
 \bottomrule \\
     & & \madry & $\mathbf{0.020}$ & $\mathbf{0.006}$ \\
     & & \wong & $0.040$ & $0.014$ \\
   MNIST & $\epsilon_2=0.5$ & CROWN-IBP & $0.059$ &  $0.027$\\
     & & IBP & $0.068$ & $0.032$ \\
     & & FAST-IBP & $0.044$ & $0.022$ \\
     \midrule
&  & \madry & $\mathbf{0.035}$ & $\mathbf{0.006}$ \\
& & \wong &  $0.100$ & $0.030$ \\
MNIST & $\epsilon_2=0.75$ & CROWN-IBP & $0.125$ & $0.060$ \\
 & & IBP & $0.145$  & $0.070$ \\
 & & FAST-IBP & $0.124$ & $0.053$\\
\midrule

& & \madry & $\mathbf{0.048}$ & $\mathbf{0.006}$ \\
 & & \wong & $0.193$ & $0.049$ \\
MNIST & $\epsilon_2=1.0$ & CROWN-IBP & $0.355$ & $0.192$ \\
&  & IBP & $0.539$ & $0.432$ \\
 & & FAST-IBP & $0.361$ & $0.175$ \\
\midrule

 & & \madry & $\mathbf{0.092}$ & $\mathbf{0.005}$ \\
 & & \wong & $0.302$ & $0.074$ \\
MNIST & $\epsilon_2=1.25$ & CROWN-IBP & $0.760$ & $0.696$ \\
 & & IBP & $0.806$ & $0.794$ \\
 & & FAST-IBP & $0.724$ & $0.652$ \\
 \bottomrule
\end{tabular}
\end{sc}
\end{small}
\end{adjustbox}
\end{center}

\caption{Results on MNIST for both $\ell_2$-ball and $\ell_\infty$-ball threat models.}
\label{table:mnist}
\end{table}
\clearpage

\paragraph{CIFAR-10} We observe that for both $\ell_\infty$-ball and $\ell_2$-ball threat models, \madry consistently achieves the best standard and robust error, with the only exception being $\epsilon=8/255$, where FAST-IBP slightly outperforms \madry in terms of robust error.

\begin{table}[h]
\begin{center}
\begin{adjustbox}{width=0.75\textwidth}

\begin{small}
\begin{sc}
\begin{tabular}{lcccc}
\toprule
  Dataset & Perturbation budget& Method & Robust error & Standard error   \\
\midrule
    & & \madry & $\mathbf{0.296}$ & $\mathbf{0.150}$ \\
     & & \wong & $0.372$ & $0.309$ \\
   CIFAR-10 & $\epsilon_\infty=1/255$ & CROWN-IBP & $0.425$ &  $0.331$\\
     & & IBP & $0.459$ & $0.356$ \\
     & & FAST-IBP & $0.384$ & $0.290$ \\
     \midrule
     &  & \madry & $\mathbf{0.378}$ & $\mathbf{0.166}$ \\
& & \wong &  $0.427$ & $0.336$ \\
CIFAR-10 & $\epsilon_\infty=2/255$ & CROWN-IBP & $0.498$ & $0.363$ \\
 & & IBP & $0.526$  & $0.415$ \\
 & & FAST-IBP & $0.476$ & $0.357$\\
\midrule
&  & \madry & $\mathbf{0.506}$ & $\mathbf{0.193}$ \\
& & \wong &  $0.609$ & $0.502$ \\
CIFAR-10 & $\epsilon_\infty=4/255$ & CROWN-IBP & $0.640$ & $0.548$ \\
 & & IBP & $0.614$  & $0.487$ \\
 & & FAST-IBP & $0.585$ & $0.449$\\
\midrule

& & \madry & $0.698$ & $\mathbf{0.239}$ \\
 & & \wong & $0.775$ & $0.729$ \\
CIFAR-10 & $\epsilon_\infty=8/255$ & CROWN-IBP & $0.673$ & $0.539$ \\
&  & IBP & $0.708$ & $0.606$ \\
 & & FAST-IBP & $\mathbf{0.675}$ & $0.546$ \\
 \bottomrule \\
     & & \madry & $\mathbf{0.217}$ & $\mathbf{0.151}$ \\
     & & \wong & $0.233$ & $0.219$ \\
   CIFAR-10 & $\epsilon_2=9/255$ & CROWN-IBP & $0.521$ &  $0.491$\\
     & & IBP & $0.457$ & $0.424$ \\
     & & FAST-IBP & $0.661$ & $0.647$ \\
     \midrule
&  & \madry & $\mathbf{0.246}$ & $\mathbf{0.151}$ \\
& & \wong &  $0.316$ & $0.287$ \\
CIFAR-10 & $\epsilon_2=18/255$ & CROWN-IBP & $0.721$ & $0.690$ \\
 & & IBP & $0.566$  & $0.515$ \\
 & & FAST-IBP & $0.763$ & $0.753$\\
\midrule

& & \madry & $\mathbf{0.298}$ & $\mathbf{0.152}$ \\
 & & \wong & $0.447$ & $0.399$ \\
CIFAR-10 & $\epsilon_2=36/255$ & CROWN-IBP & $0.780$ & $0.740$ \\
&  & IBP & $0.680$ & $0.622$ \\
 & & FAST-IBP & $0.849$ & $0.844$ \\
\midrule

 & & \madry & $\mathbf{0.386}$ & $\mathbf{0.172}$ \\
 & & \wong & $0.630$ & $0.574$ \\
CIFAR-10 & $\epsilon_2=72/255$ & CROWN-IBP & $0.900$ & $0.900$ \\
 & & IBP & $0.772$ & $0.740$ \\
 & & FAST-IBP & $0.900$ & $0.900$ \\
 \bottomrule
\end{tabular}
\end{sc}
\end{small}
\end{adjustbox}
\end{center}

\caption{Results on CIFAR-10 for both $\ell_2$-ball and $\ell_\infty$-ball threat models}
\label{table:cifar}
\end{table}

\clearpage

 \section{Additional factor ablations}

For the ablations on the perturbation budget and the margin, we focused on \(\ell_2\)-ball perturbations in~\Cref{sec:fac2,sec:fac3,sec:neurons}, as the phenomena we aim to investigate are more prominent in this context. We present similar results for the \(\ell_\infty\)-ball threat model. Additionally, while our alignment ablations in~\Cref{sec:fac1} focused on CIFAR-10 for clarity, we will now illustrate similar trends for MNIST.

\subsection{\texorpdfstring{$\ell_\infty$-ball}{linf-ball} threat model on concentric spheres} 
\label{apx:linf_ablations}
\begin{figure}[h]
    \centering
    \begin{subfigure}{0.48\textwidth}
        \includegraphics[width=\linewidth]{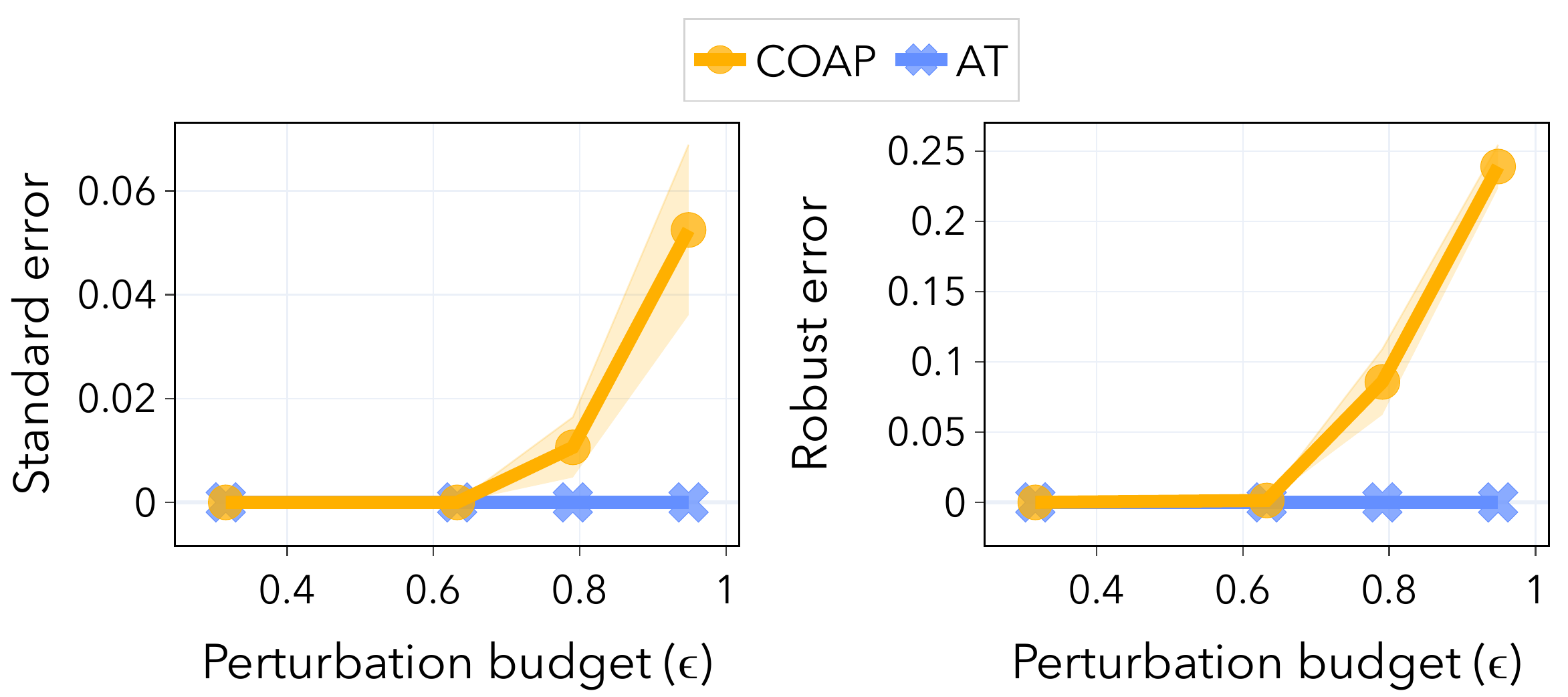}
        \caption{}
       \label{fig:spheres_eps_linf}
    \end{subfigure}%
    \hfill
    \begin{subfigure}{0.48\textwidth}
        \includegraphics[width=\linewidth]{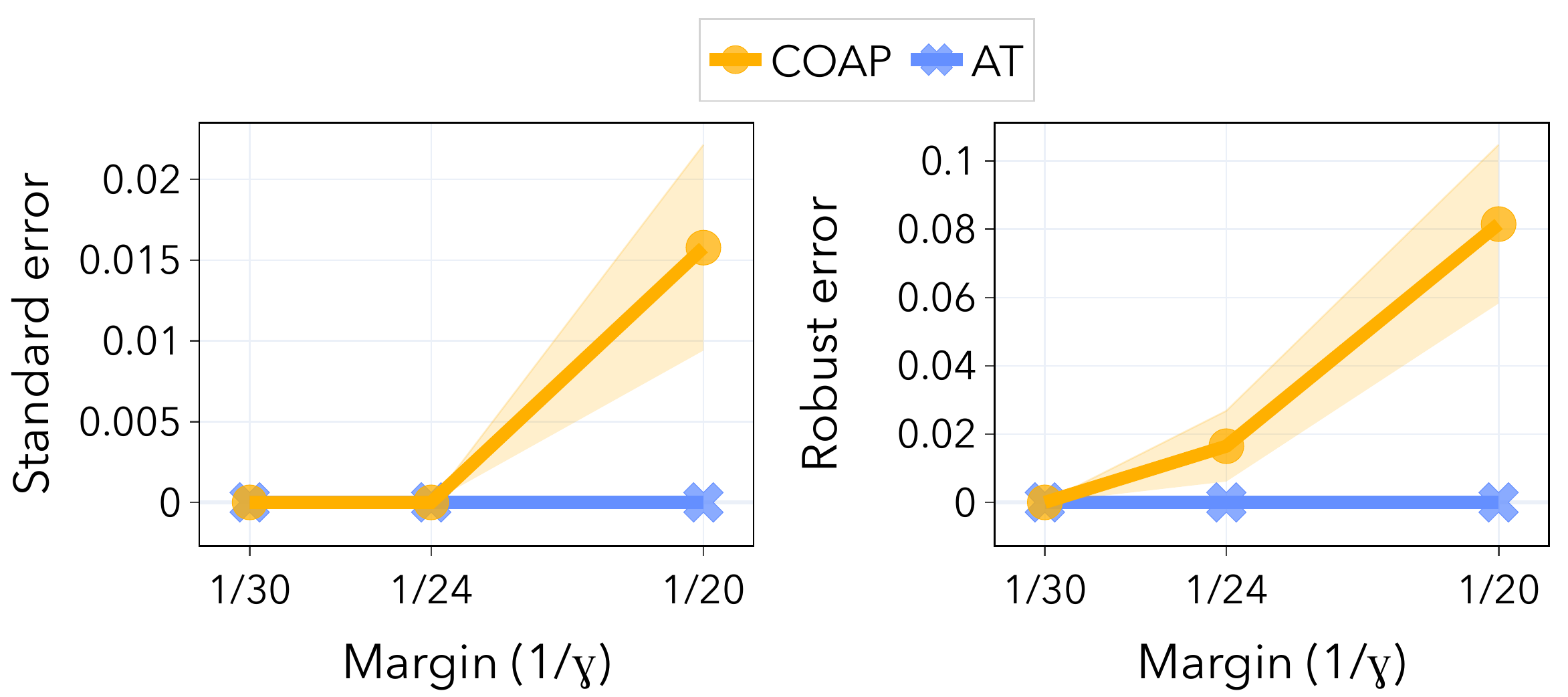}
        \caption{}
        \label{fig:spheres_gamma_linf}
    \end{subfigure}
    \caption{Results for $\ell_\infty$-ball threat model on the concentric spheres dataset. (a) Standard and robust error for \wong and \madry as the perturbation budget $\epsilon$ of the threat model increases ($\gamma=20, n=500, d=10$). (b) Standard and robust error for \wong and \madry as the inverse margin increases ($\epsilon \approx 1, n=500, d=10$).}
\end{figure}

First, we observe in~\Cref{fig:spheres_eps_linf,fig:spheres_gamma_linf} that for the concentric spheres distribution, the standard and robust error gap increases with the perturbation budget and the inverse margin.

\begin{figure}[h]
    \centering
    \hfill
    \begin{subfigure}{0.48\linewidth}
    \centering
    \includegraphics[scale=0.3]{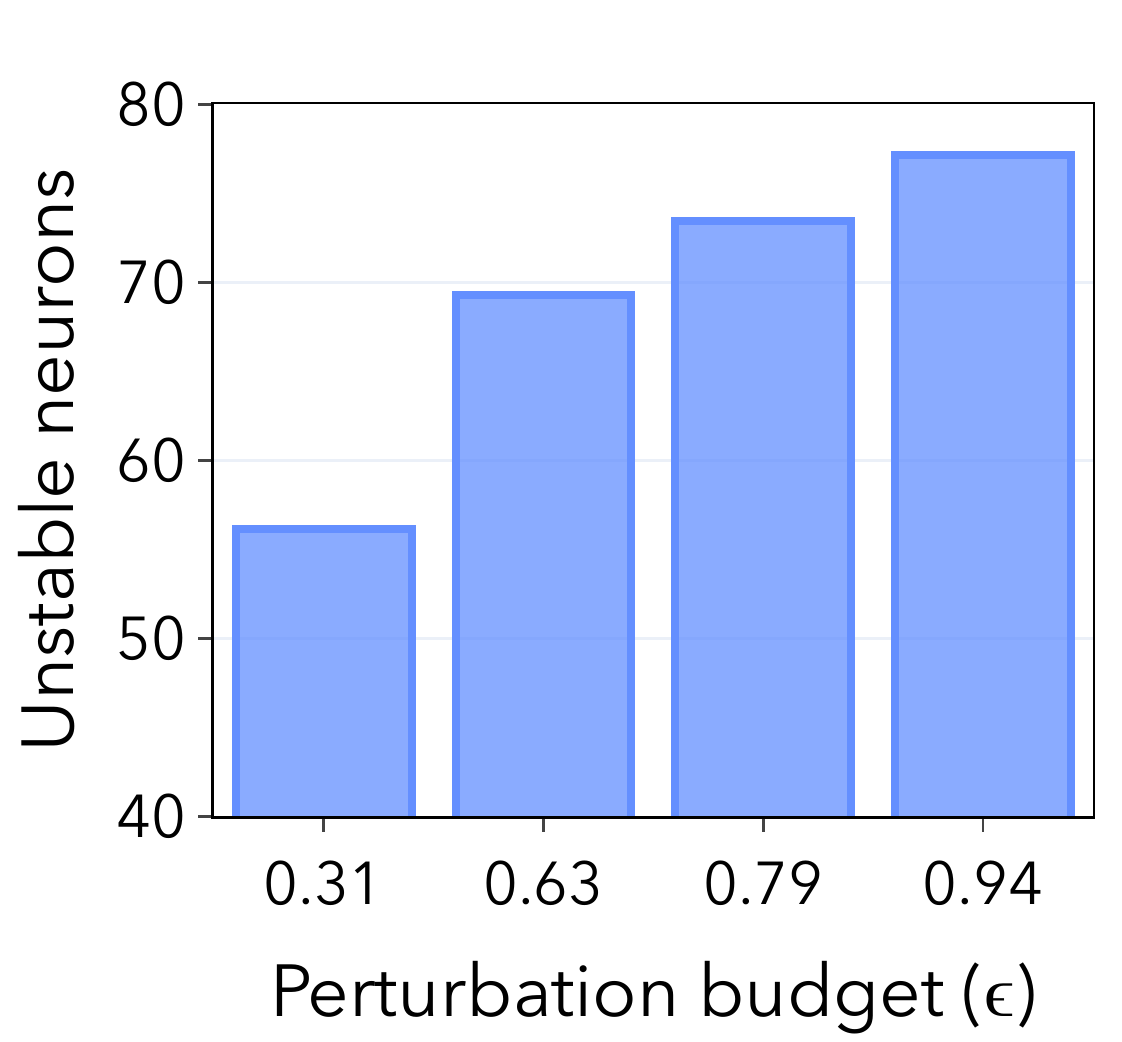}
    \caption{}
    \label{fig:spheres_un_eps_linf}
    \end{subfigure}
        \begin{subfigure}{0.48\linewidth}
    \centering
    \includegraphics[scale=0.3]{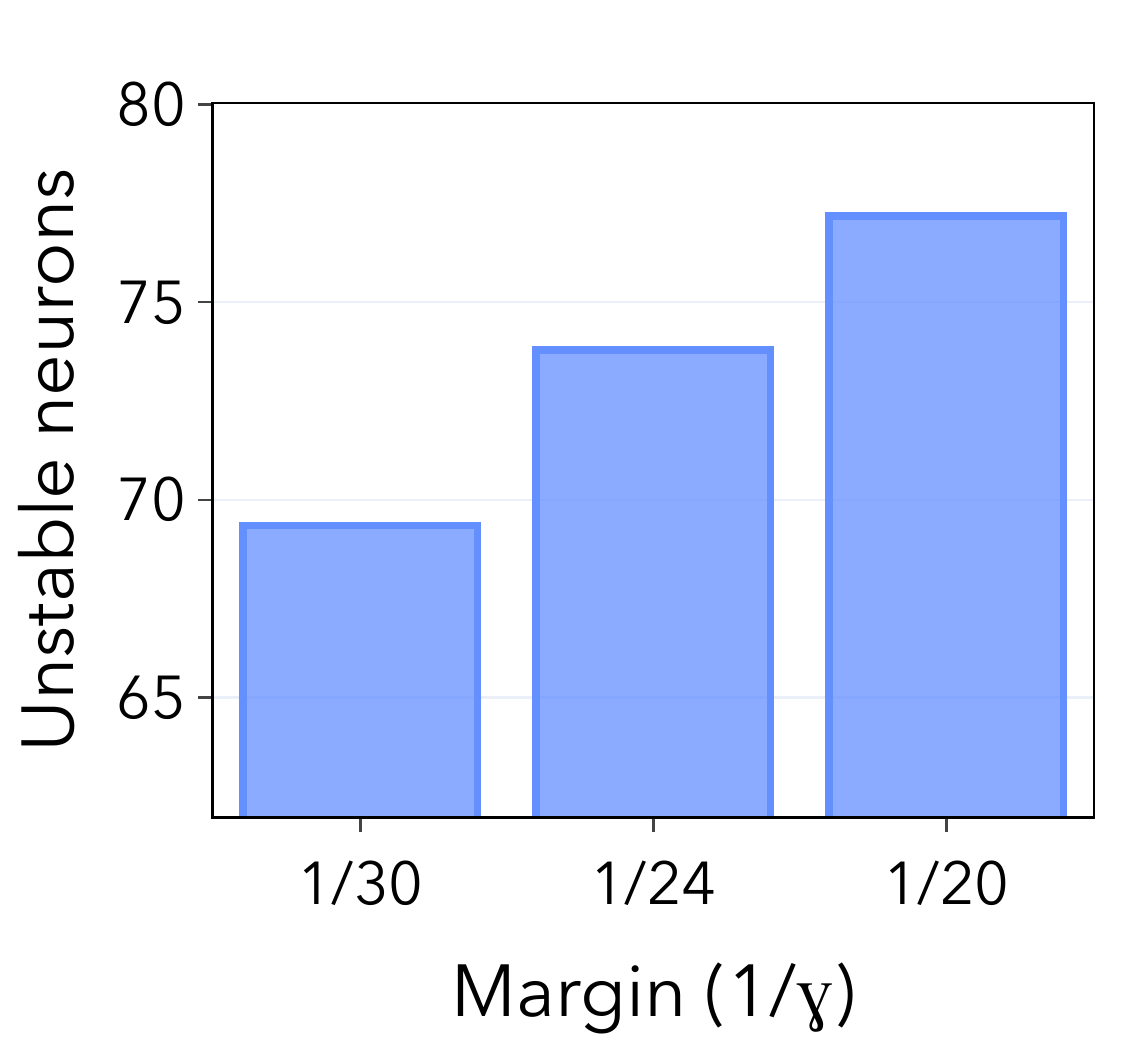}
    \caption{}
    \label{fig:spheres_gamma_vs_un_linf}
    \end{subfigure}%
    \caption{Ablations on perturbation budget and margin for the $\ell_\infty$-ball threat model. (a) Results on concentric spheres dataset ($\gamma=20, n=500, d=10$). (b) Results  on concentric spheres dataset ($\epsilon \approx 1, n=500, d=10$).}
    \label{fig:sphere_results}
\end{figure}

Further, we verify that unstable neurons, i.e. neurons with uncertain activation states given different input perturbations, are similarly affected by the factors as for the $\ell_2$-ball threat model. In particular, \Cref{fig:spheres_un_eps_linf,fig:spheres_gamma_vs_un_linf} shows that unstable neurons increase steadily with the perturbation budget and the inverse margin.

\clearpage
\subsection{\texorpdfstring{$\ell_2$-ball}{l2-ball} and signal aligned threat models on MNIST}
\label{apx:mnist_ablations}
 In addition to the ablation study conducted with CIFAR-10, presented in~\Cref{sec:fac1}, we conduct similar ablations using MNIST. 
 
\begin{figure}[th]
    \centering
    \begin{subfigure}{0.48\textwidth}
        \includegraphics[width=\linewidth]{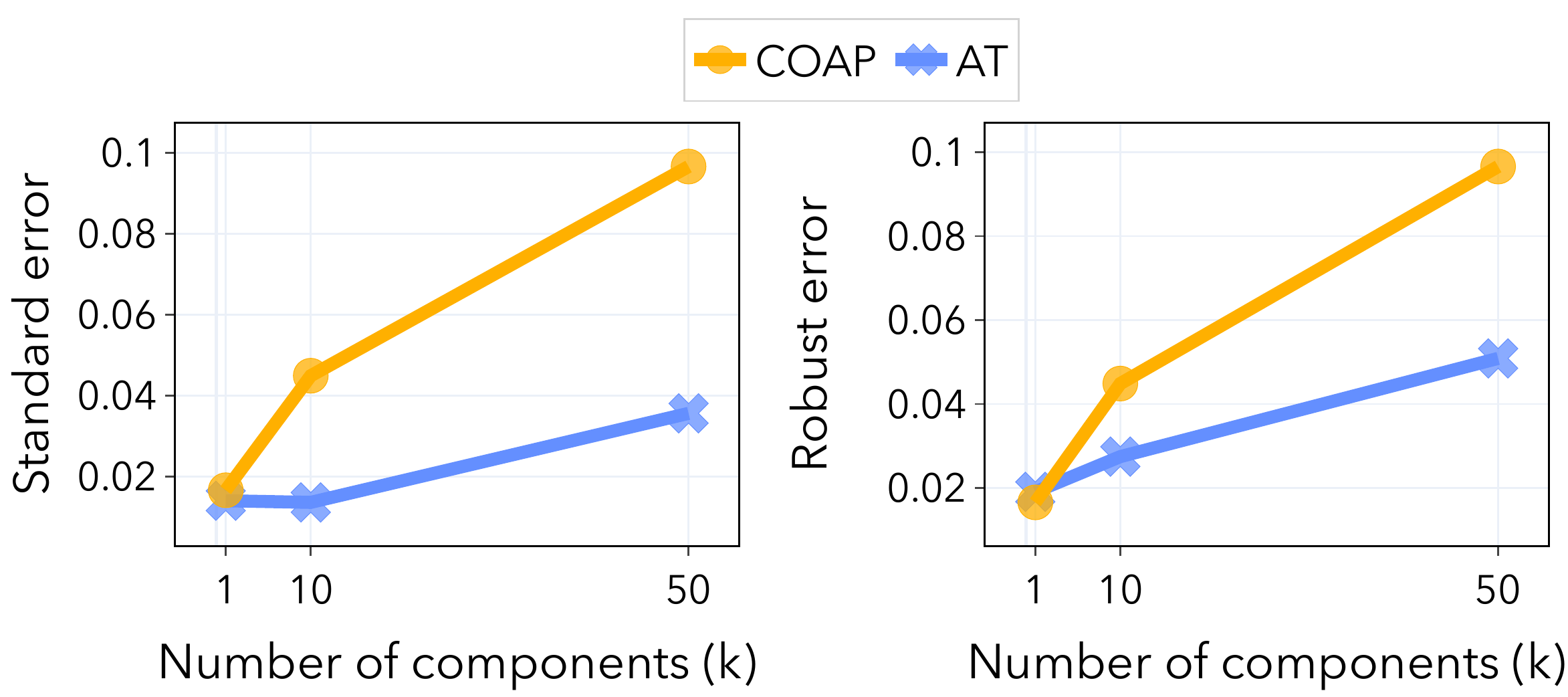}
        \caption{}
       \label{fig:mnist_k_vs_error}
    \end{subfigure}%
    \hfill
    \begin{subfigure}{0.48\textwidth}
        \includegraphics[width=\linewidth]{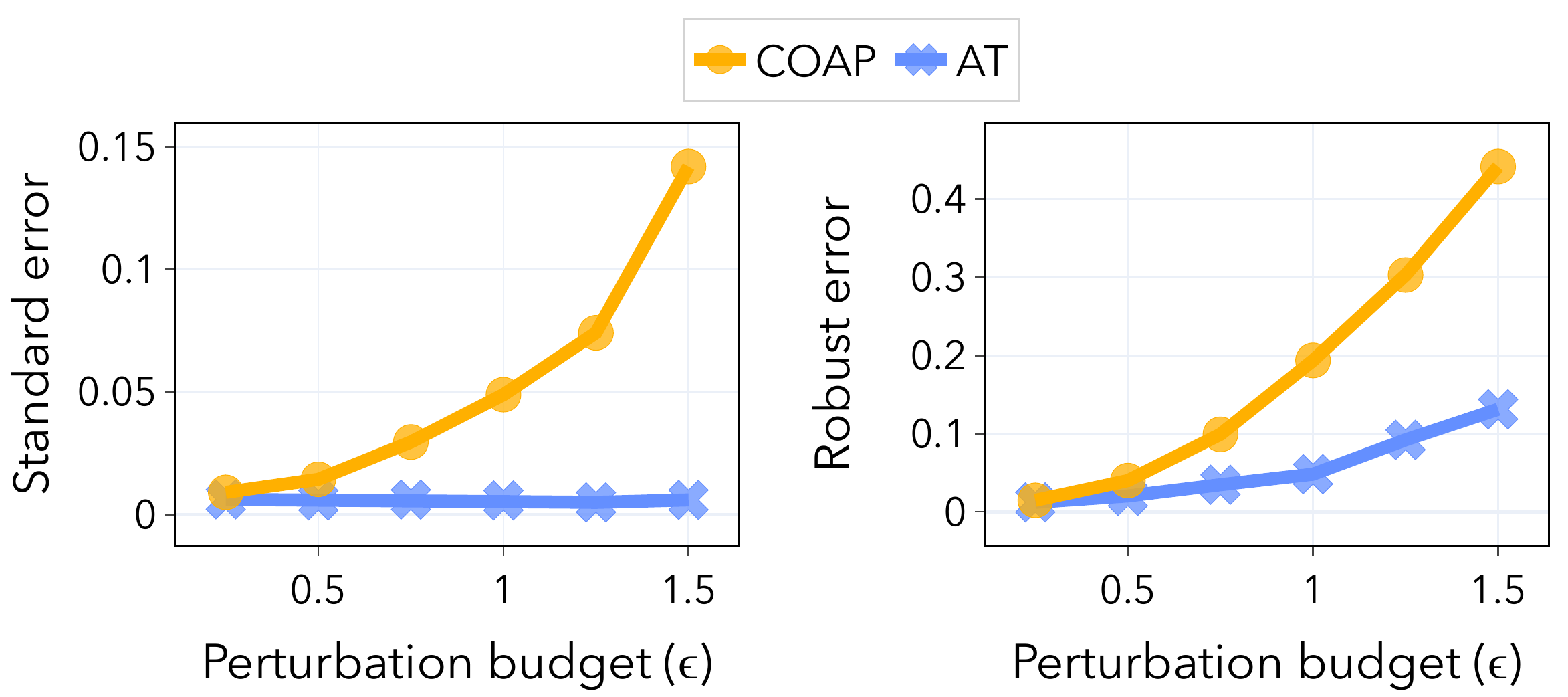}
        \caption{}
        \label{fig:mnist_eps}
    \end{subfigure}%
    \caption{Results for $\ell_2$-ball and signal aligned threat models on MNIST. (a) Standard and robust error for \wong and \madry ($\epsilon=20$) as the number of components $k$ of the threat model increases. (b) Standard and robust error for \wong and \madry as the perturbation budget $\epsilon$ increases.}
\end{figure}

 The results presented here align well with the trends identified on CIFAR-10, reinforcing our findings. In particular, we observe in~\Cref{fig:mnist_k_vs_error,fig:mnist_eps} that the standard and robust error gap increases with the perturbation budget and the alignment.

Further, we verify that unstable neurons are similarly affected by the factors as for CIFAR-10. In particular, \Cref{fig:spheres_un_eps_linf,fig:spheres_gamma_vs_un_linf} shows that unstable neurons increase steadily with the alignment and the perturbation budget.

\begin{figure}[h]
    \centering
    \begin{subfigure}{0.48\linewidth}
    \centering
    \includegraphics[scale=0.3]{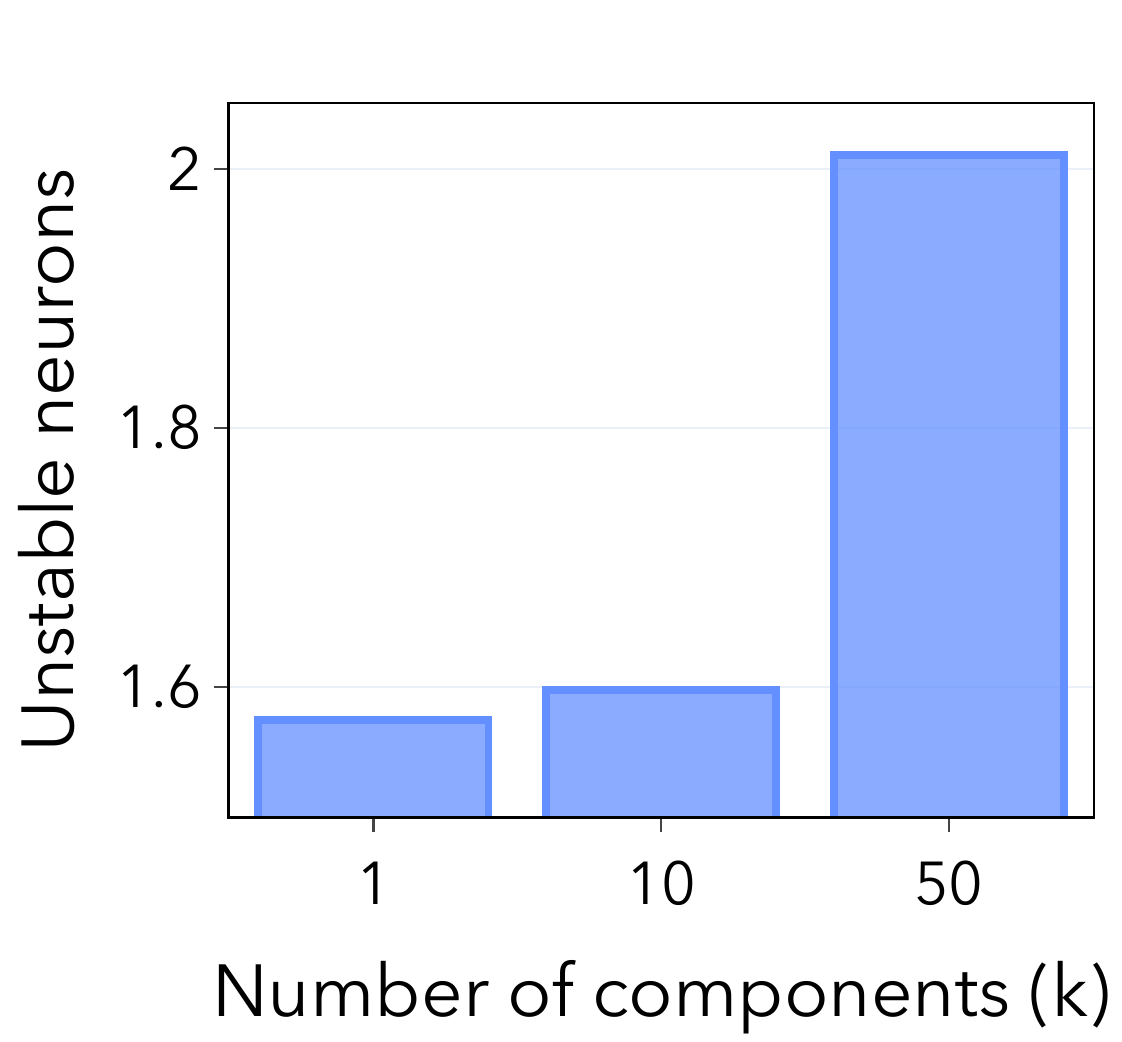}
    \caption{}
    \label{fig:mnist_k_vs_un}
    \end{subfigure}%
    \hfill
    \begin{subfigure}{0.48\linewidth}
    \centering
    \includegraphics[scale=0.3]{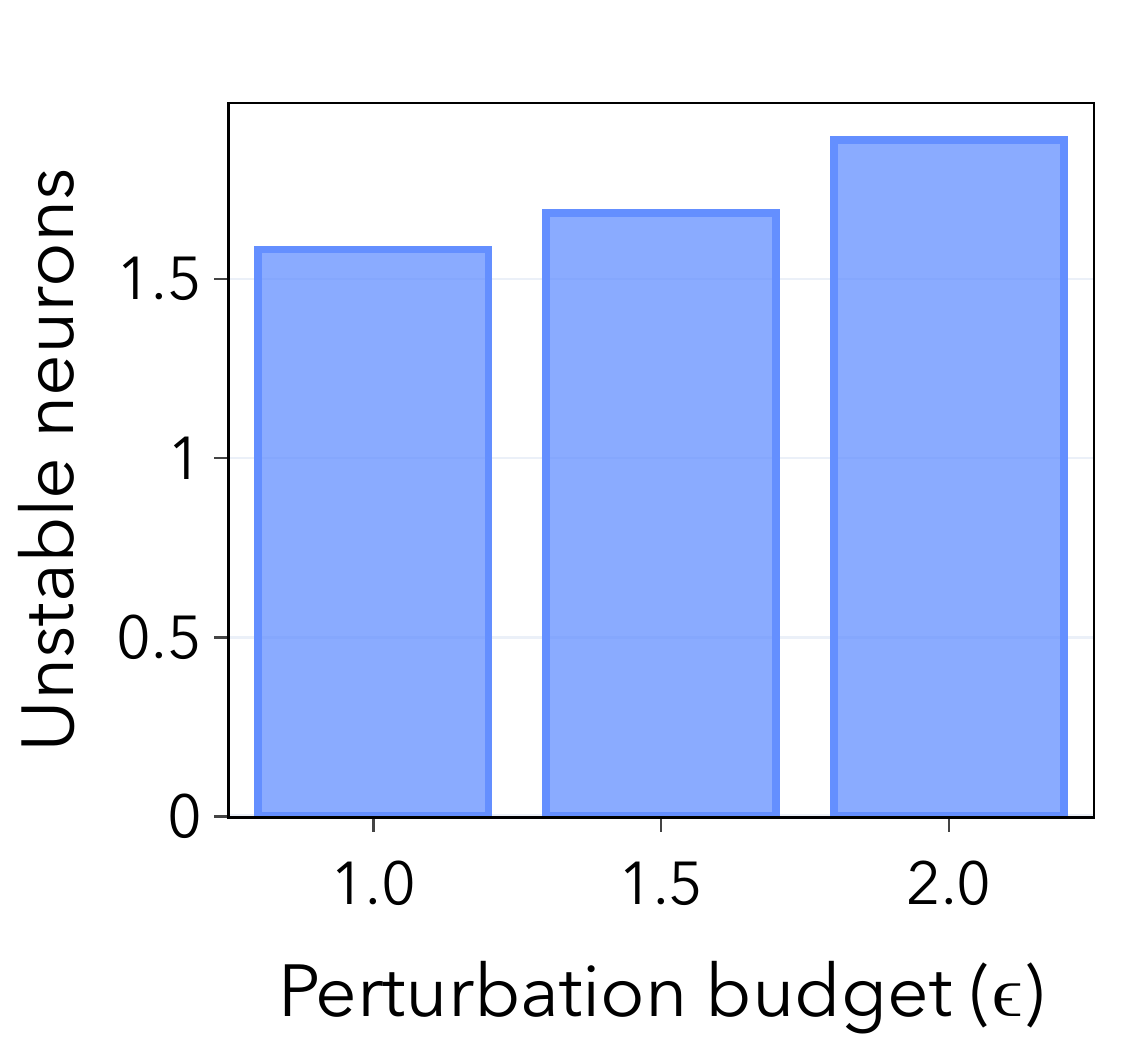}
    \caption{}
    \label{fig:mnist_un_vs_eps}
    \end{subfigure}
    \caption{Ablations for alignment and perturbation budget on MNIST. (a) Results for the signal aligned threat model ($\epsilon=20$). (b) Results for the $\ell_2$-ball threat model.}
\end{figure}

 \label{apx:mnist_exp}
 \clearpage

\end{document}